\documentclass{article}
\usepackage{preamble}  

\title{
Machine-assisted quantitizing designs: augmenting humanities and social sciences with artificial intelligence
\vspace{-0.18cm}
}
\author{
Andres Karjus \\
\normalsize School of Humanities, Tallinn University; Estonian Business School 
}
\date{\vspace{-0.8cm}}

\twocolumn
\begin{document}

\maketitle
\noindent %

\renewcommand{\contentsname}{\vspace{0.1cm}}
\renewcommand{\cftsecfont}{\normalfont}
\renewcommand{\cftsecpagefont}{\normalfont} %
\setlength{\cftsecindent}{0pt}%
\setlength{\cftsubsecindent}{0pt}%
\setcounter{tocdepth}{2}

\subsection*{Abstract}
The increasing capacities of large language models (LLMs) have been shown to present an unprecedented opportunity to scale up data analytics in the humanities and social sciences, by automating complex qualitative tasks otherwise typically carried out by human researchers. While numerous benchmarking studies have assessed the analytic prowess of LLMs, there is less focus on operationalizing this capacity for inference and hypothesis testing. Addressing this challenge, a systematic framework is argued for here, building on mixed methods quantitizing and converting design principles, and feature analysis from linguistics, to transparently integrate human expertise and machine scalability. Replicability and statistical robustness are discussed, including how to incorporate machine annotator error rates in subsequent inference. The approach is discussed and demonstrated in over a dozen LLM-assisted case studies, covering 9 diverse languages, multiple disciplines and tasks, including analysis of themes, stances, ideas, and genre compositions; linguistic and semantic annotation, interviews, text mining and event cause inference in noisy historical data, literary social network construction, metadata imputation, and multimodal visual cultural analytics. Using hypothesis-driven topic classification instead of "distant reading" is discussed. The replications among the experiments also illustrate how tasks previously requiring protracted team effort or complex computational pipelines can now be accomplished by an LLM-assisted scholar in a fraction of the time. Importantly, the approach is not intended to replace, but to augment and scale researcher expertise and analytic practices. With these opportunities in sight, qualitative skills and the ability to pose insightful questions have arguably never been more critical.

\section{Introduction}

Developments in generative large language models (LLMs, sometimes dubbed AI) have broadened their applicability to various research tasks. Of particular interest to the humanities and social sciences (H\&SS) is the capacity to use them as on-demand classifiers and inference engines. 

\tableofcontents
\onecolumn

Classifying texts or images for various properties has been available via supervised machine learning (ML) for a while. Yet, the necessity to train such models (or tune pretrained models) for every single variable of interest on sufficiently large sets of already labeled examples may have been one factor hampering the wider adoption of ML tools as research instruments (in humanities but also text-involved
industries, especially in low-resource scenarios).
Unsupervised ML, e.g. word embeddings and latent topic modeling has allowed for explorative approaches but often requires complex text preprocessing and convoluted pipelines to use for confirmatory inference. Their latent spaces are also typically opaque to interpret.

In-context (zero-shot) learning, now widely available via generative instructable LLMs, doesn't require the laborious labeling of training data. Their instruction-driven usage naturally lends itself to distilling researcher expertise into scalable big data 
annotators.
Many current LLMs, when properly instructed or tuned, can perform on par with human analysts, and occasionally even better than non-expert assistants
\parencite{webb_emergent_2023,ziems_can_2023,gilardi_chatgpt_2023,tornberg_chatgpt-4_2023,pilny_manual_2024,de_paoli_performing_2023,rathje_gpt_2024}.
There are however several challenges to such "collaborative intelligence" \parencite{mollick_co-intelligence_2024,schleiger_collaborative_2023,gonzalez-bailon_asymmetric_2023,brinkmann_machine_2023}. 
One is validation and quality control: an LLM with a high average on a public benchmark cannot be blindly expected to do well in a specific research task and context. 
There is operationalization and transparency: variables generated via classification usually constitute quantitative data, necessitating principled inference pipelines and statistical modeling to ensure the reliability of subsequent interpretations and claims.
There is little discussion yet in the recent benchmarking-centered literature on how to best operationalize the potentially very large resulting datasets of extracted variables for hypothesis testing while controlling for machine error rates.

Combining automated classification and statistics is common in some H\&SS fields, yet underutilized in others. 
Qualitative approaches or quantification without systematic testing are both nonviable when it comes to big data. This contribution takes a step beyond the performance testing focus of preceding LLM literature in H\&SS and suggests a framework for machine-assisted research, with explicated steps of systematic unitization, qualitative annotation (by humans or machines), and quantitative inference.
Arguments are presented as to why it may be a more practical alternative compared to past approaches like content analysis, and exemplary case studies are conducted to demonstrate
its application in tandem with current LLMs as instructable assistants in a range of tasks.

\subsection{Related zero-shot LLM applicability research}

This contribution builds upon and complements other recent work in what could be summarized as LLM applicability research. This is to denote the exploration of the feasibility and performance of pre-trained LLMs as research and analytics tools --- as distinct from the ML domain of LLM engineering.
ML and NLP-supported research in the (digital) humanities and (computational) social sciences is nothing new. 
However, only until recently, a typical machine-assisted text-focused research scenario would have involved either training a supervised learning classifier \parencite[or fine-tuning a pretrained LLM, see e.g.][]{majumder_interview_2020,de_la_rosa_alberti_2023} on a large set of annotated examples for a given task, or using output vectors from a word or sentence embedding LLM like BERT for clustering or other tasks \parencite[e.g.][]{fonteyn_varying_2021,sen_insider_2023}. 

Recent advances in instruction-tuned LLM technology make a difference: it has become feasible to use them for on-demand data classification and annotation,
making multi-variable machine-assisted research designs that much more feasible.
In a zero-shot scenario, a generative model is "instructed" to generate output given an input (prompt) including in some form both the analysis instructions and a unit of data. The generated outputs can be then parsed and quantified as necessary.
This obviates the need for laborious annotation work to create large training or tuning sets for every specific task, question or variable --- of which there are typically many in research dealing with complex linguistic, cultural or societal topics. Zero-shot does not necessarily outperform fine-tuning or bespoke architectures, even of smaller older models \parencite{ollion_chatgpt_2023,ziems_can_2023}, but can be significantly easier to implement for the aforementioned reason. 
Tuning or adaptation approaches \parencite{hu_lora_2021,dettmers_qlora_2023} can still be more efficient on larger or repeated tasks.

Another contributing factor to recent LLM popularity, both as chatbots and classifiers, is arguably accessibility. Running very large LLMs requires significant hardware, while various cloud services, including that of OpenAI's GPT (generative pre-trained transformer model) application programming interfaces (APIs), have made them accessible, albeit at a cost, to those who do not have access to such hardware or the skills to operate it.
To be clear though: the following discussion pertains to LLMs (e.g. GPT-4, Llama-3), but not LLM-driven chatbot services (e.g. ChatGPT, Copilot), which are not well suited for systematic data analysis.
All of this has attracted attention across various research communities well beyond NLP. Interested parties include the humanities and social sciences, which makes sense: LLMs are getting good at text-to-text tasks, and analysis in these fields often involves converting or translating complex text (in the broad semiotic sense) into some form of discrete textual codes, annotations, or taxonomies for further analysis.
In a large benchmarking exercise involving several (English-language) tasks drawn from multiple disciplines, \textcite{ziems_can_2023} show how several LLMs can achieve acceptable across various annotation and classification benchmarks. 
\textcite{gilardi_chatgpt_2023} compared the performance of the then-current GPT-3.5 to crowdsourced workers from the Amazon Mechanical Turk platform on four text classification tasks and found that it outperformed crowdworkers on accuracy and reliability (already in early 2023), while running a fraction of the crowdsourcing costs 
\parencite[see also][]{tornberg_chatgpt-4_2023,huang_is_2023,wu_llms_2023}.
Others, to name a few, have focused on performance-testing on tasks in domains like
discourse annotation \parencite{de_paoli_performing_2023,fan_uncovering_2023,rytting_towards_2023},
event data coding \parencite{overos_coding_2024},
constructions grammar and frame semantics \parencite{torrent_copilots_2023},
metalinguistic and reasoning abilities \parencite{begus_large_2023,chi_modeling_2024},
zero-shot translation \parencite{tanzer_benchmark_2024},
medical and psychology research applications \parencite{wang_text_2023,demszky_using_2023,palaniyappan_conversational_2023},
text retrieval and analytics \parencite{zhu_chatgpt_2023},
sentiment, stance, bias and affiliation detection \parencite{tornberg_chatgpt-4_2023,nadi_sentiment_2024,zhang_sentiment_2024,wen-yi_automate_2024}, including applications to languages beyond just English \parencite{mets_automated_2024,buscemi_chatgpt_2024,rathje_gpt_2024}.
There is also the artificial cognition strand interested in comparing machine and human behavior \parencite{futrell_neural_2019,taylor_artificial_2021,acerbi_large_2023,binz_using_2023}.

\subsection{Related feature analytic and mixed methods approaches}

This contribution advocates for a general-purpose quantitizing-type mixed methods research design (henceforth, QD), consisting of converting each systematically unitized data point (e.g., sentence) into a fixed number of predetermined categorical or numerical variables, followed by statistical modeling of the inferred variables (which is common in some disciplines but not widely so in H\&SS).
The terminological variability surrounding this approach necessitates a brief overview.
"Quantitization" or "quantizing" \parencite[][]{fetters_achieving_2013,hesse-biber_mixed_2010,sandelowski_quantitizing_2009} is a useful term to distinguish the process of annotating data with the explicit purpose of subsequent quantification --- from annotation for any other purposes.
"Coding" is also frequently used, but unhelpfully also has many other meanings.
This qualitative conversion step may range from quick annotation to in-depth critical or discourse analysis of every unit of data \parencite{fofana_applying_2020,banha_quantitizing_2022}, and can be quite labor-intensive.
Importantly, the research question is answered primarily based on the results of the quantification step (e.g. a statistical model), and not the data annotation step, nor by simply looking at the coded variables (as is common in some quasi-quantifying fields).

Combining qualitative and quantitative, it might be useful to situate it as mixed methods, although the majority of the latter is mixed mostly in the sense of using multiple data types (and therefore a method for each), e.g. the sequential, concurrent, convergent, triangulating and other designs \parencite{hesse-biber_mixed_2010,tashakkori_sage_2010,ocathain_three_2010,huynh_critical_2019}.
Variants similar to QD have also been referred to as
"integrated" \parencite{tashakkori_sage_2010,creamer_introduction_2018,ohalloran_interpreting_2019},
"integration through data transformation" design \parencite[not to be confused with "transformative mixed methods";][]{mertens_transformative_2008},
"qualitative/quantitative" \parencite{young_online_2013}, 
and "converting" \parencite{creamer_introduction_2018}. 
Qualitative comparative analysis exemplifies an extreme transformation approach that involves reducing (even potentially very complex) data into binary truth table variables \parencite{vis_comparative_2012,kane_mixed_2023}.
\textcite{parks_natural_2023} discuss a "dialogical" machine-assisted framework which is not a quantitizing approach but also uses NLP tools.
"Distant reading" in digital humanities \parencite{moretti_distant_2013} and "ousiometrics" \parencite{fudolig_decomposition_2023} are also machine-assisted approaches, but typically rely on counts of words or topic clusters and not manual coding.

A mixed quantitizing approach where the annotation step is referred to as "coding" is found within content analysis \parencite{schreier_qualitative_2012,banha_quantitizing_2022,pilny_manual_2024}. However, subsequent quantification of the coded variables is not considered as "a defining criterion" of CA \parencite{krippendorff_content_2019}, which also includes interpretative-qualitative approaches \parencite{hsieh_three_2005}, and quantification limited to counting or limited statistical testing \parencite[cf.][]{morgan_qualitative_1993,schreier_qualitative_2012}.
Limited quantification is 
occasionally found in discourse analysis \parencite[e.g.][]{ohalloran_interpreting_2019}.
Political event coding also applies coding in this sense \parencite{schrodt_automated_2013,overos_coding_2024}.
Thematic analysis does as well but typically lacks
statistical modeling of the resulting distributions \parencite[cf.][]{braun_thematic_2012,trahan_toward_2013}, which is a problem (see Methods). This includes recent LLM-powered proposals \parencite{de_paoli_performing_2023,hau_towards_2024}.

Issues with quantification, aggregation, and biased sampling affect rigor and replicability and can lead to spurious results \parencite{parks_natural_2023}.
Approaches like that --- incorporating quantification in a limited manner, either by using impressionistic claims ("more", "less", "some") without actual quantification, or counting but stopping before proper statistical or otherwise suitable estimation --- will be referred to as quasi-quantifying, going forward.
This contribution emphasizes the need for systematic statistical modeling, to be able to estimate uncertainty and deal with confounding variables, interactions, multicollinearity, and repeated measures (more common in the humanities than often assumed). These are often dismissed in e.g. small-scale humanities research, but cannot be ignored when applying automation to operationalize large datasets.

One domain where the approach of combining qualitative coding with subsequent quantification is widespread, if not the default, is usage-based (corpus-based, variationist, cognitive) linguistics, with roots in componential analysis and structural semantics \parencite{goodenough_componential_1956,nida_componential_2015}.
Often enough it is not referred to as a specific design but has been called "usage feature analysis", \parencite[sometimes prepended with "multi-factorial";][]{glynn_quantitative_2010}, or instead "behavioral profiles" \parencite{gries_corpus-based_2009}.
The quantitizing, often also called "coding", is typically conducted by the researchers themselves, as it often requires expert linguistics knowledge. As for coding schemes, sometimes also called "classification schemes", standard variables like grammatical categories from past literature can be used \parencite{szmrecsanyi_culturally_2014}, but schemes may also be developed for a given research question \parencite[cf.][]{glynn_testing_2010}. Developing a standardized coding scheme or taxonomy for future research can also be the sole aim, as in branches of psychology, where similar methods are used \parencite[][]{hennessy_developing_2016}.
Unlike some of the disciplines mentioned above, a great deal of attention is paid to rigorous statistical procedures in the quantitative modeling step \parencite{wolk_dative_2013,gries_most_2015,winter_statistics_2020}.

\subsection{A machine-assisted quantitizing design (MAQD)}

\begin{figure}[htb]
	\noindent
	\includegraphics[width=\columnwidth]{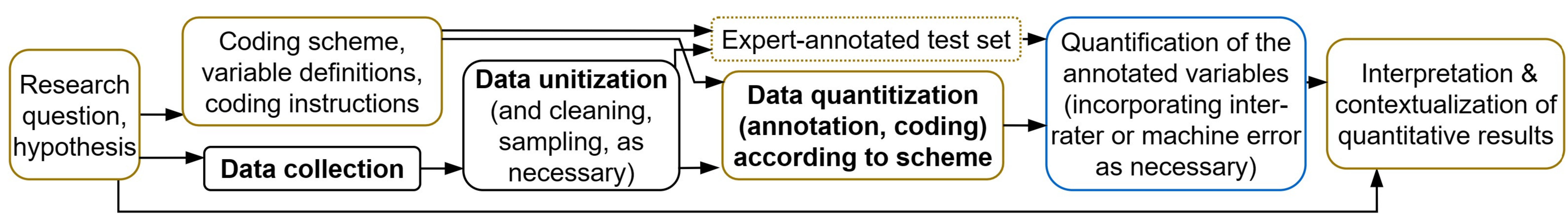}
	\caption{A typical QD pipeline. Qualitative elements are outlined in yellow, quantitative in blue. Steps where machine assistance (ML, LLMs, or otherwise) may be applied are in bold, including the quantitization step. Annotating a smaller additional test set is optional but strongly recommended if using either multiple human annotators or a machine.
	}\label{fig_pipeline}
\end{figure}

The framework discussed in the following is just that: the QD or feature analytic design, generalized beyond linguistics, with augmentation by capable machines to solve the scaling problem, and emphasizing the need for good practices in unitizing, agreement or error estimation, and principled statistical analysis.
The suggestion is to think as machines such as LLMs as tools, or in a sense (narrow) artificial intelligence assistants, to scale expert analysis to larger datasets --- but not as "oracles" or "arbiters" in the taxonomy of \textcite{messeri_artificial_2024}. The final quantification-driven interpretation is left to the human researcher.
It is suitable where the data is qualitative (text, images, etc) but can be quantitized (annotated, coded, converted) into one or more categorical or numeric variables for quantitative modeling.
A typical pipeline can be summarized as follows, further illustrated in Figure \ref{fig_pipeline}; see Methods and the Supplementary Information (SI) for details on these steps.

\begin{enumerate}[nosep] 
    \item Research question, hypothesis or explorative goal; data collection.
    \item Coding scheme with application instructions; unitization principles.
    \item Unitization of data (into meaningful units of analysis); sampling and filtering if necessary.
    \item Qualitative annotation (quantitizing) of each unit according to a coding scheme of one or more variables. If delegated to machine, then also: acquiring or annotating a test set.
    \item Quantitative (typically statistical) analysis of the inferred variables, their relationships, and uncertainty. Incorporation of the rates of (human) disagreement or (machine) error in the uncertainty estimation.
    \item Qualitative interpretation of the quantitative modeling (potentially in combination with examples from data or theory).
\end{enumerate}

\subsection{What this contribution does}

As discussed above, most of this pipeline is already conveniently standard to many. Applying machine learning is nothing new in some fields of H\&SS, although it is usually not conceptualized as mixed or quantitizing methods. This is a practical proposal for the wider adoption of one tried and tested design well suited for maximizing gains from machine automation. It is hoped to be helpful as a framework that would be easily referenceable, implementable, replicable, and teachable. It can function as an end-to-end analytic framework, or, being theory-agnostic, integrated into existing methodologies such as content or thematic analysis.
This contribution has three goals: filling one research gap and contributing to two others.
It describes and encourages the adoption of the QD, given its comparative advantages over alternatives.
The Methods section positions the approach in the context of other common designs and describes its components, the importance of statistical modeling and confidence estimation, and how to incorporate machine (or human) error rates in the former. The general QD can be applied using just human annotators.
However, as shown in the case studies below, there are benefits to automation, currently in particular via generative LLMs as on-demand annotators or retrieval tools, yielding scalability to magnitudes of data that might be unfeasible in purely qualitative or human-annotation paradigms.
Size alone should never be the goal in science, but larger data (if properly sampled) can lead to more representative findings and greater statistical power to effectively model the complexities of human language, culture, and societies.
Secondly, the supporting case studies exemplify the application of MAQD principles, while also complementing current LLM applicability research in H\&SS which has been mostly focused on large languages like modern English or clean contemporary data or both. The case studies include 9 languages: Estonian, Finnish, German, Italian, Japanese, Latin, Russian, Turkish, and English in four varieties (19th century US, 18th century UK, contemporary standard US, and US American as used on social media). 
Thirdly, the case studies illustrate applying the MAQD and complement related benchmarking literature discipline-wise, with experiments covering several H\&SS domains --- linguistics, sentiment and discourse analysis, literature and translation analysis, media and film studies, history, social network science, lexicography, and discussing applications of multi-modal models in visual cultural analytics. 
The examples include replications of past research and synthetic examples created for this contribution. 
While some exemplify the MAQD pipeline, others include practical tasks like LLM-driven data cleaning and content filtering, which are frequent prerequisites to applying any QD type approach, especially if the data source is a larger general sample such as a corpus.
Unlike most related research \parencite{ziems_can_2023,rathje_gpt_2024}, the focus here is intentionally not on public benchmarks or shared tasks. 
One reason is data contamination \parencite{aiyappa_can_2023}, that current LLMs trained likely on open internet data may well include publicly available datasets.
In the one NLP benchmark utilized here (one highly unlikely to cause contamination), the zero-shot approach scores 1.5-2x above the state of the art.
Secondly, relying on big public benchmarks would be misaligned with the proposed framework, which advocates for smaller but expert-annotated, task-specific test sets --- to be used to estimate machine error rates, in turn incorporated in subsequent statistical estimates; but also to compare and choose models (including fine-tuned and personal or research group specific models; see Discussion).
The code and test sets are made public, to complement the present benchmarking scene and to provide an easy starting point for interested parties.

\subsection{What this contribution is not: three disclaimers}

Some disclaimers are in order, as "artificial intelligence" has
attracted a significant uptick of public attention and corporate hype in 2023-2024. 
First, we will not be explicitly dealing here with topics like data copyright, related ethical issues, possible model biases, environmental concerns, machine "psychology", "AGI", or AI tractability and theoretical limitations. 
These issues have been and will be discussed elsewhere
\parencite[][]{bender_dangers_2021,lund_chatgpt_2023,rooij_reclaiming_2023,tomlinson_carbon_2024,liesenfeld_opening_2023,motoki_more_2023,feng_pretraining_2023,binz_using_2023,ollion_dangers_2024,novozhilova_looking_2024,hagendorff_machine_2024}.
There is also a growing literature centered around what LLMs are not or should not be capable of \parencite{asher_limits_2023,rooij_reclaiming_2023,dinh_large_2023,barone_larger_2023,sclar_minding_2023,dentella_systematic_2023}.
Here the focus is pragmatic, on using suitable machines as research tools where applicable, guided by expert evaluation. 

Secondly: this is not about replacing researchers or research assistants \parencite[cf.][]{erscoi_pygmalion_2023}, but about augmenting, capacitating, and empowering, while promoting transparent and replicable research practices.
Human labor does not scale well, machines do; human time is valuable, machine time is cheap. This is about reducing repetitive labor, leaving more time for meaningful work.
\textcite{ziems_can_2023} suggest that "LLMs can augment but not entirely replace the traditional [computational social science] research pipeline." Indeed, the largest gains are likely to be made from empowering expert humans with powerful machine annotators and assistants,
not assembly-line automation of science \parencite{lu_ai_2024}.
This proposal places qualitative thinking and expert knowledge front and center, requiring designing 
meaningful hypotheses, coding schemes, and analysis instructions, expert annotation of evaluation data sets, and final contextualization of the results of the statistical step.

Finally: the LLM test results and classification accuracies reported in the case studies should only be seen as the \emph{absolute minimum baseline}. 
Model comparison or prompt optimization is not the goal here; most prompts consist of simple 1-2 sentence instructions (see SI). 
Detailed expert prompts tend to yield better results. 
The accuracy rates are therefore not the focus, although they are reported --- to illustrate tasks with (already present) potential for automation and to discuss how to incorporate them in subsequent statistical modeling.

\section{Method details and statistical considerations}

This section aims to situate and summarize some crucial aspects of applying a machine-assisted design. Practical aspects such as data preparation, data unitization, 
setting up an example LLM for machine annotation, and further statistical and transparent science considerations are all expanded upon in the SI, due to space constraints. 
See also \textcite{tornberg_best_2024} for further advice on setting up LLM annotators.

\subsection{Conceptual method comparison}

As an inherently mixed methods framework, MAQD incorporates advantageous aspects of qualitative and quantitative designs, while overcoming their 
respective limitations. 
A point-by-point method comparison can be found in the SI. The following simplified summary is intended to situate the MAQD, but naturally, these archetypes may not capture every research scenario. Qualitative designs are deeply focused and can consider wider context, reception, societal implications, power relationships, and self-reflection, but the analyses are often difficult to generalize and estimate the uncertainty of, hard to replicate or scale to large data.
Quasi-quantifying designs try to retain this depth, and systematic coding supports replicability, but variable relationships and uncertainty remain impressionistic, while overconfidence in quantitative results without statistical modeling can lead to spurious results (see SI).
Primarily quantitative methods are scalable (but only given countable data), relationships and their uncertainty can be estimated, and replication (or full reproduction, given data and procedures) is easier. Yet they may be seen as lacking the nuance and depth of the above.
QD like feature analysis combines the benefits of qualitative analysis in the coding or transformation step with systematic quantification and uncertainty estimation. While coding involves subjectivity, it can be replicated, and statistical analyses reproduced. It can be applied in exploratory and confirmatory designs regardless of discipline. Yet big data is a challenge due to the laborious human-annotation bottleneck.
Machine assistance enables scaling up the QD, while retaining all the benefits of qualitative analysis, systematic quantification, and replicability. 

Even without machine assistance, the QD provides a more systematic framework compared to approaches limited in these aspects. 
This includes applications to "small" datasets like interviews or fieldwork data.
In addition to open data, also coding schemes, data processing, and statistical analysis code or instructions should be published whenever possible to foster transparent scrutiny and substantive rather than conceptual or personal critique.
QD is also a good alternative for getting at the  "bigger picture" or themes, like in thematic analysis: unitizing and subsequent systematic quantitative modeling of theme distributions can only improve the induction of overarching themes, and unlike quasi-quantifying practices, also helps control for confounds, etc.
It can also supplement discourse or content analysis research or approaches building on social semiotics \parencite{halliday_language_1978}.
While the focus here is academic, the same approach could be used in analytics in business, marketing, media monitoring and communication, etc. \parencite[][]{dellacqua_navigating_2023,mets_automated_2024}.

\subsection{Necessity of statistical modeling to avoid unintentional quasi-quantifying designs}

The QD and MAQD only make sense if the inferred variables are subsequently modeled within a statistical framework to estimate the uncertainty of the estimates, whether they involve prevalence counts or complex multivariate relationships between the variables.
In a hypothesis-driven research scenario, this typically also entails accounting for possible confounding variables, interactions, and often enough, repeated measures.
None of these issues are exclusive to quantitative research, yet unfortunately often ignored in qualitative designs. 
There are some approaches like qualitative comparative analysis that promise the evaluation of interactions, but only after significant reduction of data complexity \parencite{kane_mixed_2023}.

Estimating the effect size (though impossible in qualitative designs) of a difference, trend or association further helps to avoid making sweeping yet spurious claims based on an effect that may actually describe only a small portion of variance in a (typically complex) social or cultural system with many interacting variables.
This also makes simple pairwise tests an often insufficient solution.
More versatile models like multiple regression enable estimating the uncertainty of the main hypothesis while controlling for confounds, interactions, and in the mixed effects (multilevel) case, repeated measures \parencite[cf.][]{gries_most_2015,clark_should_2015,winter_statistics_2020,mcelreath_statistical_2020}.

Inherent repeated measures are common in H\&SS. Survey and interview data typically contain multiple responses, often with a variable number from multiple respondents. Literary or artistic examples may be sourced from multiple works from multiple authors from multiple further groupings, like eras or collections. Linguistic examples may be sourced from corpora (with several underlying sources) or elicited from multiple informants. Social media data often contains multiple data points per user. If the underlying grouping or hierarchical structure of a dataset is not accounted for, estimates can reverse direction, known as Simpson's paradox \parencite[][]{kievit_simpsons_2013}; see the SI for examples. This issue also applies to qualitative designs, but it is impossible to systematically control for repeated measures there.

A necessary prerequisite for reliable analysis of initially unstructured or qualitative data (such as free-running text, film, etc.) is principled unitization, partitioning the data into meaningful divisions or units. In some cases the units are fairly obvious, e.g. paintings in a study about art, or sentences in syntax research; in others less so. It is important to make sure the units are comparable, large enough to be meaningful yet small enough to be succinctly analyzable (see the SI for further discussion). 
In summary, taking a systematic approach to both the initial data as well as the inferred discrete variables is crucial to applying a QD or MAQD. 
There is not enough space here to delve into all possible pitfalls (and plenty of handbooks do), 
but a lack of control over these issues can easily lead to false, overestimated conclusions or even diametrically opposite results. 
Any quantitative claim, regardless of the label on the design, should be accompanied by an estimate of confidence or uncertainty. 
If claims about a population (in the statistical sense) are made based on a sample, the reliability or replicability of any claimed differences, tendencies, prevalences, etc. should be estimated. The smaller the samples the more so, to avoid claims based on what may be mere sampling noise. 
In a crowdsourcing or machine-assisted scenario, quantifying annotation error is equally important.
These issues are however not something that should be seen as a complicating factor or a requirement to make the researcher's life harder. On the contrary, it makes it easier: instead of worrying if an analysis is replicable and representative, the uncertainty can be estimated, enabling more principled final interpretation and decision-making.

\subsection{Incorporating classification error in statistical modeling}

In a quantitizing research design, regardless if the annotation step is completed by humans or machines, inter-rater (dis)agreement should be accounted for in any subsequent operationalization of these new data, to avoid overconfident estimates and making Type I errors. It is far from atypical for (also human) annotation tasks to have only moderate agreement. This aspect is often ignored, even in QD applications like linguistic usage feature analysis, which usually otherwise strives for statistical rigor. 
As discussed in the Introduction, no methodological element in this proposal is new on its own, including that of using LLMs to annotate data, zero-shot or otherwise.
What appears to be not yet widespread is the suggestion to systematically use expert knowledge to delegate coding, analysis, or annotation tasks to machines such as LLMs, while --- importantly --- also making sure the machine error rates are incorporated in statistical modeling and uncertainty estimates. 
Doing so enables using less than a perfect 100\% accurate classifiers or annotators without fear of making overconfident or biased inferences down the line.
Unless a closely comparable test set already exists, this will typically require a subset of data to be manually coded by human annotator(s) for evaluating the chosen machine(s).

Annotation error can be accounted for in several ways, including errors-in-variables (EIV) type regression models (if the variables are numerical), directly modeling measurement errors using an MCMC or Bayesian approach \parencite{carroll_measurement_2006,goldstein_modelling_2008}, or using prevalence estimation techniques \parencite[often referred to as "quantification" in machine learning literature;][]{gonzalez_why_2017}. Distributional shift naturally remains problematic but can be mitigated to an extent \parencite{guillory_predicting_2021}.

\begin{figure}[htb]
	\noindent
	\includegraphics[width=\columnwidth]{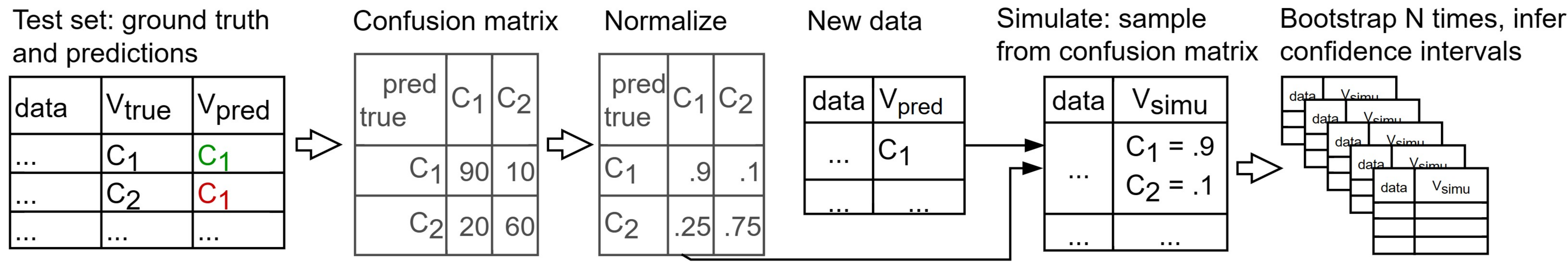}
	\caption{A bootstrapping-driven pipeline for estimating the uncertainty in a machine-annotated categorical data variable $V$. The crucial component is the test set for comparing human expert annotation (ground truth) and machine predictions (or that of human coders). This provides an estimate of annotator accuracy and class confusion within the variable, which can then be used in bootstrapping the confidence intervals for the statistic of interest.
	}\label{fig_bootstrap}
\end{figure}

Keeping it simple, a bootstrapping approach is considered here which applies straightforwardly to exploratory and confirmatory scenarios and makes use of the rich error information available via computing the confusion matrix between a ground truth test set predictions and predictions or annotations. 
The procedure is simple, involving simulating the annotation procedure by sampling from the confusion matrix (see Figure \ref{fig_bootstrap} for illustration; ground truth is rows, and predictions are columns).
\begin{enumerate}[nosep]
    \item Compute the confusion matrix $m$, of machine vs ground truth, for the variable of interest $V$ which has a set of categorical levels or classes $C$
    \item For each class $i \in C$, normalize the count distribution of predictions against ground truth ("rows" of $m$) as probability distributions $d_i$
    \item Perform bootstrapping, creating $N$ number of synthetic replicates of the data, where each predicted value $V_j$ is replaced with a simulated value
        \begin{enumerate}[label*=\arabic*.,nosep]
        \item For each case $V_j$ with a value $C_i$, perform random weighted sampling from $C$ using the corresponding $d_i$ as weights
        \item After simulating all values of $V$, perform the statistical analysis of interest (counts, prevalence, regression, etc.) on this synthetic dataset, and record the output(s)
        \end{enumerate}
    \item Calculate $\pm$ confidence intervals for the statistic(s) of interest based on the estimate of the error (e.g. standard deviation) in the bootstrapped outputs.
\end{enumerate}

For example, if the goal is to estimate the confidence of a percentage of class $C_i$ in $V$, then the process is to perform bootstrapping on the raw new (classified or annotated) data some large number of times (e.g. 10000), by sampling from the test set confusion matrix for each case in $V$, and calculating the statistic (percentage) in each replicate. Finally, calculate 95\% confidence intervals via e.g. normal approximation ($1.96\cdot\sigma$).
The intuition is: if the outputs of the (human or machine) annotator match with the designated ground truth 100\%, then there will be no variance in the sampling either, and intervals will be $\pm0$. The more confusion in the confusion matrix, the more variance in the replicates, leading to higher error estimates or wider confidence intervals.
This is the simplest approach, and potentially better and more robust procedures may be considered.

\section{Results of case studies}

For a class of machines to be considered for automation and augmentation in a machine-assisted design, their outputs should align closely enough with what human annotators or analysts would produce,
while the remaining disagreement or error should be quantified and taken into account. Instructable LLMs have been shown to perform well in many domains and tasks, although often limited to English and clean data. This section exemplifies the application of LLM-assisted MAQD through case studies in several languages and levels of difficulty as discussed in the Introduction.
Examples include explorative tasks, emulated analyses based on synthetic data, and a handful of replications of published research. Most cases, being somewhat simplified examples, involve coding schemes of 1-2 variables, while real QD applications in complex cultural or social topics would typically involve more variables (in turn necessitating multivariable statistical modeling, see Methods).
Two cloud service LLMs were used,
\mbox{gpt-3.5-turbo-0613} and \mbox{gpt-4-0613} \parencite{openai_gpt-4_2023}, current at the time of conducting the studies, referred to as GPT-3.5 and GPT-4. The last, visual analytics example was updated in revision to use the newer \mbox{gpt-4o-2024-08-06} instead.
Accuracy and Cohen's kappa are used as evaluation metrics of machine performance in most case studies (see Table 1 in the SI for a bird's eye view of the scores), as the focus here is agreement
with human-annotated ground truth, rather than retrieval metrics like F1. Accuracy illustrates empirical performance, while kappa adjusts for task difficulty by considering the observed and expected agreement given the number of classes.

Most of the case studies emulate or replicate only a segment of a research project pipeline, 
focusing on the applicability of LLMs in the quantitization step, but QD components of unitization, coding scheme, and (potential) quantification are described in each example. Further details on some case studies are found in the SI, along with the LLM prompts, due to space limitations.
Many examples boil down to text-to-text, multi-class classification tasks --- as much of H\&SS is also concerned with various forms of classification, interpretation, and taxonomies, to enable predictions and discovering connections.
Most tasks do not yield 100\% agreement between human and machine annotations and analyses. This is not unexpected --- less so because these machines have room to improve, and more because these are mostly qualitative tasks requiring some degree of subjective judgment, where expecting 100\% human agreement would not be realistic either. Human agreement rates could indeed be used to estimate an upper bound for machine accuracy in a given task.

\FloatBarrier

\subsection{Example case: topic classification instead of latent topic modeling}

In fields like digital humanities and computational social science, topic modeling is a commonly used tool to discover themes, topics, and their historical trends in media and literary texts, or to conduct "distant reading" \parencite{moretti_distant_2013,janicke_close_2015}. The bag-of-words LDA \parencite{blei_latent_2003} is still commonly used, as well as more recent sentence embedding-driven methods \parencite[][]{angelov_top2vec_2020,grootendorst_bertopic_2022}. 
These are all forms of soft or hard clustering: good for exploration, but suboptimal for confirmatory research. Often, historical or humanities researchers may have hypotheses in mind, but they are difficult to test when they need to be aligned to ephemeral latent topics or clusters. Instead of such a tea leaves reading exercise, one could instead classify texts by predefined topics of interest. Until now, this would have however required training (or tuning an LLM as) a classifier based on labeled training data. Annotating the latter is time-consuming and easy out-of-the-box methods like LDA have remained attractive \parencite[cf.][]{jelodar_latent_2019,jacobs_topic_2019,sherstinova_topic_2022}.

With instructable LLMs, laborious training data compilation is no longer necessary, and topics can be predicted, instead of derived via clustering. Good prompt engineering is still necessary, but this is where qualitative scholars can be expected to shine the brightest.
\textcite{ziems_can_2023} worry that topic modeling may be challenging for transformer-based language models, given their context window size limitations, but even this is quickly becoming a non-issue. 
The GPT-4 version used here had a window size at 8000 tokens (about 6000 English words); the next iteration of GPT-4o models, released while this paper was in review, are already at 128,000, and Google's Gemini-1.5 models at 2 million \parencite{kilpatrick_gemini_2024}.
Also, longer texts can and indeed often should be split into smaller units.

The zero-shot topic classification approach is exemplified here using a dataset of Russian-language synopses of Soviet era Moscow-produced newsreels from 1945-1992 \parencite[for details see][]{oiva_framework_2024}. They summarize the story segments that comprise the roughly 10-minute weekly newsreel clips (12707 stories across 1745 reels; each on average 16 words). As part of the collaboration cited above, an expert cultural historian predetermined a set of 8 topics of interest based on preceding research and close viewing of a subset of the reels, from politics to social and lifestyle to disasters (the latter was not encountered in the dataset; details and the longer topic definitions in the SI).
An additional "miscellaneous" topic was defined in the prompt to subsume all other topics. Such an "everything else" or negative category is a naturally available feature in a zero-shot approach while requiring more complicated modeling in traditional supervised machine learning.
We used an English-language prompt despite the data being in Russian, as this yielded better accuracy in early experiments \parencite[this makes sense given the dominant language of the model; cf.][]{wendler_llamas_2024}.

To summarize this in MAQD terms: the coding scheme consists of the categorical topic variable with 9 levels, and the numeric variable of year (already present in the dataset). The unit is a story synopsis. The quantitization step consists of determining the primary topic of a given unit. The following quantification step involves aggregating topic counts as percentages, followed by regression analysis, with the machine-annotation uncertainty estimated using the aforementioned confusion matrix bootstrapping approach based on the expert test set. The final qualitative interpretation step could further contextualize the regression results with examples from the corpus, as in \textcite{oiva_framework_2024}.

Our expert annotated a test set of 100 stories for topics, one per story. Preliminary prompting experiments indicated that a single example per instruction prompt would yield the highest accuracy, which was 0.88 for GPT-3.5 (kappa 0.85; 0.84 and kappa 0.8 for GPT-4). This is of course the more expensive option when using cloud services like that of OpenAI that charge per input and output length. The cheaper option of batching multiple examples -- preceded by an instruction prompt, requesting multiple output tags --- seemed to reduce classification accuracy (this deserves further inquiry). 
While 88\% accuracy is not perfect, recall that this is on a 9-class problem and a historical dataset rife with archaic terms and abbreviations that may not exist in the training data of a present-day LLM. 
The synopses, though short, sometimes contain mentions of different themes and topics. For example, a featured tractor driver in an agricultural segment may also be lauded as a Soviet war hero or local sports champion. 

\begin{figure}[htb]
	\noindent
	\includegraphics[width=\columnwidth]{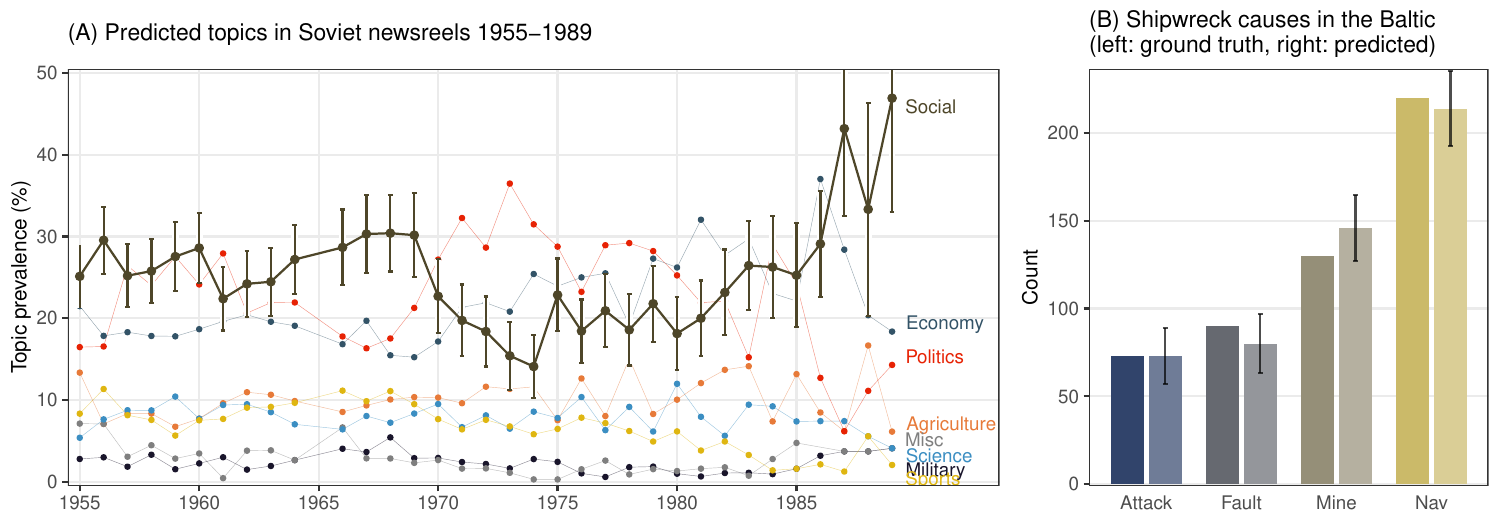}
	\caption{(A) Zero-shot prediction of predefined topics in the corpus of Soviet newsreel synopses. Vertical axis shows yearly aggregate percentages. Bootstrapped confidence intervals are added to the trend of the Social topic. There are less data in the latter years, reflected in the wider intervals.
    (B) Wrecking causes of ships found in the Baltic Sea, mostly in Estonian waters, as annotated by experts based on field notes and historical documents (left), compared to zero-shot prediction of said categories based on the same data, with bootstrapped confidence intervals on the counts. Due to fairly good classification accuracy, the counts end up roughly similar.
	}\label{fig_topics}
\end{figure}

Following testing, GPT-3.5 was applied to the rest of the corpus of 12707 stories, producing an estimate of news topic prevalence in the Soviet period \ref{fig_topics}.A. Among the trends, there is an apparent increase in the Social topic towards the end. However, the classifier is not 100\% accurate; there are also fewer issues and therefore fewer data points towards the end. To test the trend, one can fit for example a logistic regression model to the period of interest (1974-1989), predicting topic by year (binomial, Social vs everything else). The model indicates a statistically significant effect of $\beta=0.064, p<0.0001$: each passing year multiplies the odds of encountering the Social topic by a factor of $e^{0.064}=1.07$).

However, this is still based on the predicted topics. One way to incorporate this uncertainty is to use bootstrapping, as discussed in Methods: simulate the classification procedure by sampling from the confusion matrix of the test set (the annotated 100 synopses), and rerun the statistical model over and over on a large number of bootstraps. This yields distributions for each statistic of interest for inferring confidence intervals. Since the classifier here is fairly accurate, the 95\% confidence interval around the log odds estimate comes to $\pm0.02$, and for the p-value, $\pm0.00002$ (in other words, the upper bound still well below the conventional $\alpha=0.05$).
The same procedure was applied to infer intervals for the percent estimates in \ref{fig_topics}.A.

Latent topic models are useful for exploration and discovery, but this exercise shows that zero-shot topic prediction is a viable alternative for testing confirmatory hypotheses about topical trends and correlations. 
If distributions of topics as in LDA are desired, lists or hierarchies of topic tags could be generated instead.
In an additional but limited exploratory exercise, a sample of about 200 random synopses (about 8000 words in Russian, or 16k tokens) was fed into a GPT-3.5 version with a larger context window \mbox{(gpt-3.5-turbo-16k)}, prompting it to come up with either any number, and then also a fixed number of general topics. These outputs were quite similar to the one initially produced by our expert historian by their admission.

\subsection{Examples of historical event cause analysis and missing data augmentation}

Detecting and extracting events, their participants, and causes from texts, including historical documents, is of interest central to many fields in H\&SS and information retrieval \parencite{sprugnoli_novel_2019, lai_event_2021}.
\textcite{ziems_can_2023} experimented with applying LLMs to binary event classification and event argument extraction in English. 
Here, GPT-4 is applied to quantitizing the causes of shipwrecking events in an Estonian-language dataset of maritime shipwreck descriptions in the Baltic \parencite[a part of the Estonian state hydrography database][]{his_hudrograafia_2023}. 

The unit of data here is the database entry, containing
a description of an incident based on various sources and fieldwork (n=513, 1591-2006 but mostly 20th century), including a primary cause of the wrecking variable. The already present initial coding scheme consists of a semi-open categorical variable with 54 unique values, containing a term or a phrase, inferred from the longer description by experts (used as ground truth here). This was simplified as a four-category variable: direct warfare-related (assaults, torpedo hits), mines, leaks or mechanical faults, and broadly navigational issues (such as getting caught on shallows or in a storm). 
The quantitization step consisted of inferring the category from the description texts, 
which range from short statements on sinking cause and location to longer stories such as this (here translated) example: 
\textit{Perished en route from Visby to Norrköping on the rocks of Västervik in April of 1936. After beaching in Gotland, Viljandi had been repaired, set sail from Visby to Norrköping around mid-month. In a strong storm, the ship suffered damage to its rudder near Storkläppen, losing ability to steer. The injured vessel drifted onto the rocks of Sladö Island, and was abandoned. Local fisherman Ossian Johansson rescued two men from the ship in his boat. One of them was the ship's owner, Captain Sillen. Wreck sank to a depth of 12 meters.} (this is tagged as a navigation and weather-related wrecking).

GPT-4 accuracy is fairly high: the primary cause prediction matches with human annotation 88\% (kappa 0.83; but e.g. the "mine" class has a 100\% recall). This is very good for a task with often multiple interacting causes and a sometimes arbitrary primary one, 
and on texts full of domain-specific terminology and Estonian maritime abbreviations. Figure \ref{fig_topics}.B illustrates the quantification step, 
answering the question of the prevalence of wrecking causes, 
and how much a predicted distribution of causes would differ from an expert-annotated one, with bootstrapped confidence intervals again added to the counts (see Methods).

A quantization task like this can also be used or viewed as missing data augmentation or imputation, where a variable with entirely or partially missing data is populated based on other variables. Real-world datasets in H\&SS are often sparse, while applying multivariable statistics typically requires complete data.
Another similar exercise was conducted using data from a Finnish-language broadcast management dataset, attempting to infer the missing production country entries of TV shows and films based solely on their (also very short) descriptions. The task was born out of real necessity when dealing with a highly sparse dataset. Despite its complexities and very open-ended nature (further details and examples in the SI), the results were promising, with GPT-4 yielding 72\% accuracy against ground truth.
These explorations demonstrate 
that already current LLMs are quite capable of completing inference even in contextual tasks that would otherwise have required manual work by domain experts.

\subsection{Example of relevance classification with LLM-driven OCR correction in digitized newspaper corpora}
A quantitizing design often requires first filtering and extracting relevant units from a larger pool of data.
Digitization efforts of historical textual data such as newspapers and books have made large-scale, machine-assisted diachronic research feasible in many domains where this was not possible before. However, finding relevant examples from vast swathes of digitized data, more so if it is plagued by optical character recognition (OCR) errors, can be challenging.
A part of the pipeline from a recent work \parencite{kanger_deep_2022} is replicated here as an annotation exercise. The study sought to measure centurial trends in dominant ideas and practices of industrial societies focusing on the topic of nature, environment, and technology, based on digitized newspapers from multiple countries.
Their pipeline for retrieving relevant examples for further analysis consisted of assembling a set of broad keywords, extracting passages where these occur, and estimating if they match the topic ("nature" vs "human nature", etc.). The LDA topic modeling required cleaning and lemmatizing, and annotating the abstract topics for relevance. Such pipelines can be streamlined as a single operation on an LLM. 

The authors kindly provided a human-annotated set of 99 excerpts for this exercise,
extracted from a corpus using a first-pass keyword search. As such, the unit here is a context window around a keyword, the coding scheme is the categorical variable (nature-related or not), and quantitization consists of determining whether a given text snippet is nature-related. As a filtering step, there is no further quantification here, but a real application could further model the co-occurrence of nature with other topics or variables or test hypotheses about temporal trends.
While some parts of the sampled corpus were fairly readable, others contain OCR-distorted examples such as this: \\ 
\textit{principally to easing in » u ¿ allan consolidated bonds nine Issues Siorln « fallí and on'y two lssues ßalnl » 8 The lïttei Included the 3 . per cent 1942 in which laigf pa'cek were bou.ht The Syd Ii , banks lollnqulshed a small pait of recent rlim Arünstnaturalleacilon in t . limited S , r of issues the main body of Indu- irai continued to find keen support}. \\
This can however be "cleaned" with GPT-4: \textit{principally to easing in Australian consolidated bonds; nine issues showing a fall and only two issues gaining. The latter included the 3 per cent 1942, in which large parcels were bought. The Sydney banks relinquished a small part of recent gains. As a natural reaction in the limited set of issues, the main body of industrial continued to find keen support.} \\ 

While such operations may suffer from LLM hallucination issues, we can test if this step degrades or improves the downstream classification results. 
The case turns out to be the latter. The prompt was to classify a given test input as mentioning "nature or environment in the biological natural world sense, including nature tourism, landscape, agriculture, environmental policy". Without the cleaning, GPT-3.5 gets 0.79 accuracy (0.49 kappa) and GPT-4: 0.9 (0.77). With cleaning, GPT-3.5 gets 0.82 (0.56) and GPT-4: 0.92 (0.82 kappa). This is another difficult task with limited and historical period-specific contexts, and more precise prompting might help. The results are promising however \parencite[in contrast to similar work e.g.][]{boros_post-correction_2024}.
Using zero-shot or fine-tuned LLMs may well provide a simpler and faster alternative to complex processing and annotating pipelines such as this, as well as obviate the need for parameterizing and complex mathematical operations to make embedding vectors usable for that \parencite[cf.][]{sen_insider_2023}, streamlining the application of MAQD approaches.
The combination of an initial faster filter (keyword search, embedding similarity) with a follow-up generative LLM-based refining filter may well be a fruitful approach, as running an entire corpus through a cloud service LLM like GPT-4 can be costly (and time-consuming, even if using a local model).

\subsection{Linguistic usage feature analysis applications}

As discussed in the introduction, machine learning and LLMs have found various uses in branches of linguistics (beyond the explicitly labeled computational one). The application of LLMs in the usage feature analysis approach appears to be less explored. This case study replicates a part of the pipeline of a recent paper on linguistic value construction in 18th-century British English advertisement texts \parencite{mulder_linguistic_2022}. It focused on modifiers such as adjectives as a way of expressing appreciation in language and tested hypotheses about historical trends of both modifier usage and the advertised objects themselves.
The paper details the process of developing the categories of "evaluative" (subjective) and "descriptive" modifiers via systematic annotation exercises and normalizing spelling in the historical texts (heterogeneous by nature and plagued by OCR errors) using edit distance, word embeddings, and manual evaluation. While cleverly utilizing computational tools, it is evident that no small amount of manual effort was expended. Most of such manual work can be streamlined and automated using zero-shot LLMs, including in low-quality OCR scenarios.

The replication here starts from the annotation, where the phrase units --- such as \textit{servants stabling} or \textit{fine jewelry} --- had already been extracted from a larger corpus (this is also a potential application for LLM-powered retrieval or filtering, as in the previous section). The coding scheme is the subjectivity variable (two levels); the quantitization step consists of assessing the relative subjectivity of the adjective in the phrase. A subsequent potential quantification step could aggregate those to compare against a time period variable as in the paper.
GPT-4 agreement with expert human annotations is high (accuracy 0.94, kappa 0.89). For context, the paper reports the kappa agreement between the two researchers to have been at 0.84 in the first annotation iteration. This is not an easy task either and may require subjective decisions; e.g. \textit{servants horse} is tagged descriptive (objective) yet \textit{gentleman's saddle} as subjective in the dataset. 

The same approach could also be used to automate or augment the quantitizing step in domains like grammar and syntax \parencite[cf.][]{szmrecsanyi_culturally_2014,begus_large_2023,qin_is_2023,beuls_construction_2025}.
In another recent study \parencite{karjus_evolving_2024} we experimented with automating semantic analysis, measuring linguistic divergence in US American English across political spectra. The unit: pairs of social media posts (N=320). The coding scheme: an ordinal similarity variable from the DURel scheme \parencite{schlechtweg_diachronic_2018}; the quantitization: determine if a given target word or emoji (of 8 in total) is used in the same or different sense between them. GPT-4 achieved moderate correlation with the two human annotators ($\rho=0.45$ and $0.6$), struggling with the limited available context and context-specific emojis, underscoring the need to verify LLM accuracy against human judgment (see Methods and SI).

\subsection{Example of social network inference from literary texts}

\begin{figure}[htb]
	\noindent
	\includegraphics[width=\columnwidth]{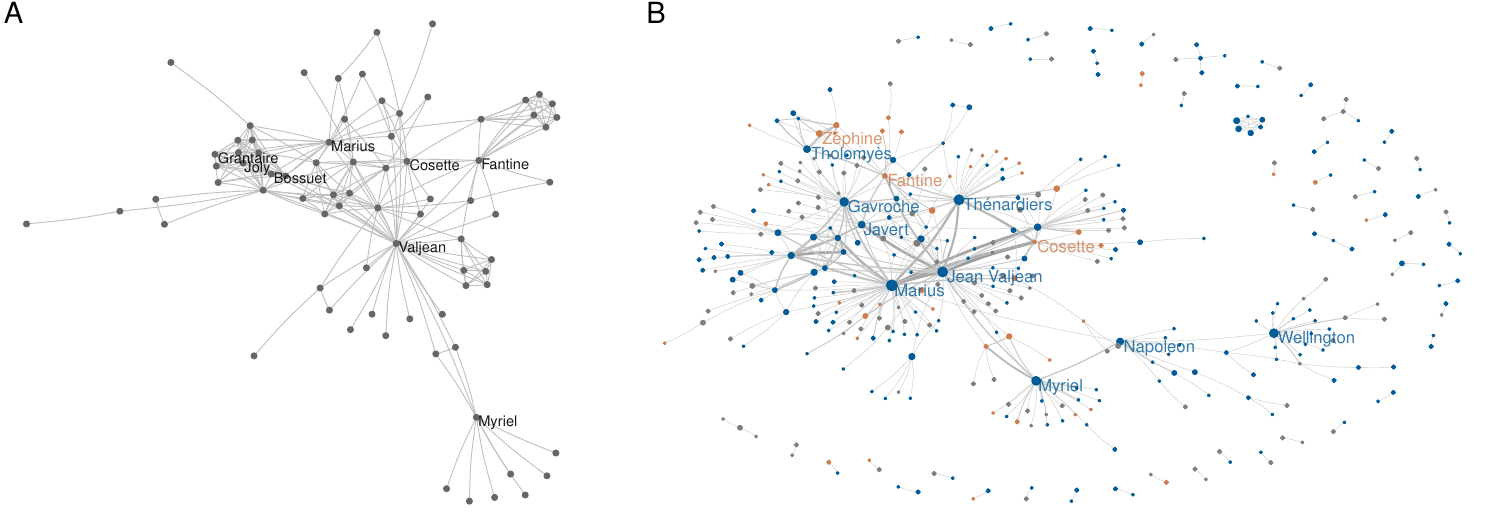}
	\caption{Social networks of interacting characters in "Les Misérables" by Victor Hugo, manually constructed textbook example on the left (A), and as automatically inferred using LLM on the right (B; men are blue and women orange).
	}\label{fig_nets}
\end{figure}

Another potential application for MAQD is quantitizing data for network analysis. Here, an entire book was analyzed for character interactions: the unit is a chapter, the coding scheme consists of an open-ended annotation variable for any persons that interact (coded as pairs), and a gender variable for each character. The quantitization step involves finding instances of interacting pairs in a chapter (and marking character gender), and quantification consists of operationalizing the list of pairs as a weighted undirected network, where the nodes are characters and the links between them are weighted by the number of recorded interactions.
Figure \ref{fig_nets}.A depicts a character network manually constructed from "Les Misérables" by Victor Hugo, often used as a textbook example in network science and related fields. \ref{fig_nets}.B is a network of interacting characters inferred from the English version of the same book 
using GPT-3.5 according to the scheme above, and GPT-4 for estimating gender.
The result may have some errors --- some anomalous pairs like street names and unspecific characters ("people" etc.) required filtering, and better results may be achieved with a newer model.
Still, the result is much richer than the smaller manual version, including non-plot characters discussed by Hugo in tangential sections. This limited exercise demonstrates the applicability of LLMs to unitizing and quantitizing entire texts (the resulting units for the subsequent MAQD quantitative step being character pairs here), without requiring preprocessing with specialized models e.g. named entity recognition, syntactic parsing \parencite[cf.][]{elson_extracting_2010}.

\subsection{Exemplary genre quantification in literature and film}

In their LLM applications paper, \textcite{ziems_can_2023} discuss the computational analysis of literary themes, settings, emotions, roles, and narratives, and benchmark some of these tasks. 
The following is a replication of a recent study \parencite{sobchuk_computational_2024} which sought to compare computational methods of capturing and clustering thematic (genre) similarity of literary texts against manually assigned genres (Detective, Fantasy, Romance, Sci-Fi), using the Adjusted Rand Index \parencite[ARI;][]{hubert_comparing_1985}. Combinations of text embedding algorithms (bag-of-words, topic models, embeddings), parameters, preprocessing steps, and distance measures were evaluated.
This replication illustrates both how instructable LLMs can be used as simpler alternatives to complex parameterized computational pipelines, and how to conceptualize longer texts such as books in a MAQD. 

Given that ARI is equivalent to Cohen's kappa \parencite{warrens_equivalence_2008}, the performance of an LLM, set up to classify genres, can be directly compared to the clustering results. The authors kindly shared their labeled 200-book test set. Rather than parsing entire books, 25 random passages 
(the unit here; 5000 in total) were sampled from each. The coding scheme consists of the genre variable, and GPT-3.5 was instructed to label each passage as one of the 4 genres.
The best-performing parameter and model combination in \textcite{sobchuk_computational_2024} used complex preprocessing, a doc2vec embedding \parencite{le_distributed_2014}, and cosine similarity for clustering. The preprocessing involved lemmatizing, NER and POS tagging for cleaning, and lexical simplification (using another word embedding). This yielded an ARI of 0.7.
The simple zero-shot LLM approach here achieved a (comparable) kappa of 0.73 (0.8 accuracy) without any preprocessing, and only judging a subset of random passages per book (each book tagged simply by majority label; more data and a better model would likely improve results).
Some genres were easier than others (Fantasy had 100\% recall), while books combining multiple genres complicate things. These results reflect the result of the first case study: instead of clustering or topic modeling, zero-shot learning enables direct prediction and classification, and LLMs often obviate processing \parencite[see also][]{chaturvedi_where_2018,sherstinova_topic_2022}. 

\begin{figure}[htb]
	\noindent
	\includegraphics[width=\columnwidth]{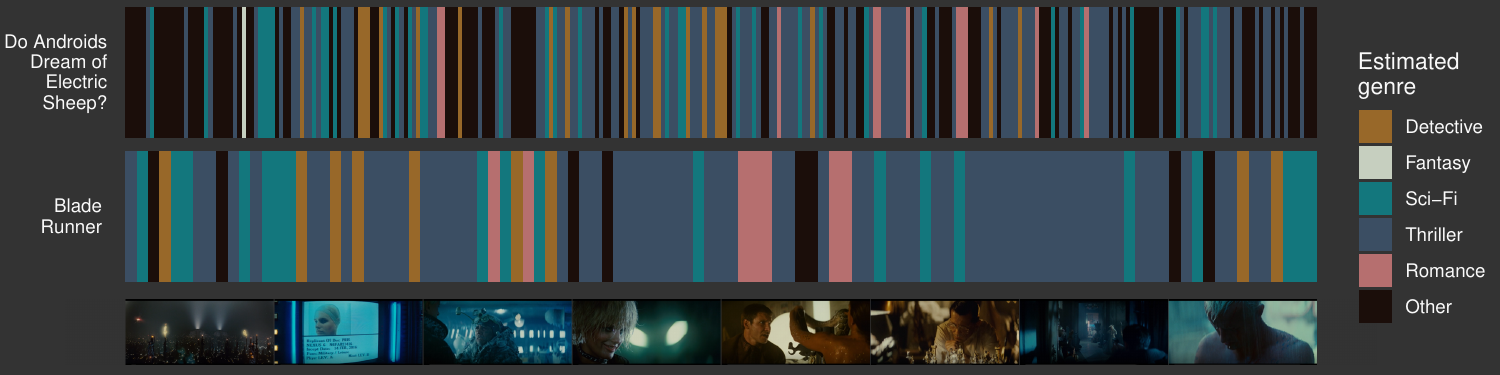}
	\caption{Zero-shot classification of genre across one book and its film adaption, split into equally-sized segments and scenes, respectively. Frames from the film are added for illustration. Differences and similarities become readily apparent, and can provide basis for follow-up qualitative or quantitative comparisons.
	}\label{fig_reels}
\end{figure}

Instead of labeling entire books with a single genre, on-demand classification like this can easily yield a more informative distribution of genres within each.
Figure \ref{fig_reels} illustrates an example application using two texts, P.K. Dick's "Do Androids Dream Of Electric Sheep?", and the adapted script of "Blade Runner" by H.~Fancher and D.~Peoples (the 1981 version). The script was unitized by scenes (but merging very short ones), and the book as equally sized chunks (note that the book is 3 times longer), and each unit classified using GTP-3.5.
Differences between the book and movie are revealed, the latter being more of a thriller with sci-fi and detective elements, while the book delves into various other topics. Both have detective elements in the first half and romantic ones in the middle. The one segment labeled "fantasy" does include the text \textit{"At last a bird which had come there to die told him where he was. He had sunk down into the tomb world."}
This illustrates the potential of using zero-shot LLMs for scene analytics and quantifying fiction, either for exploratory or statistical modeling purposes, without needing to train specialized classifiers 
for each variable, while multi-modal LLMs can also add a visual dimension. Beyond this simple example, one could also consider running time or align book and adaption \parencite[cf.][]{yi_movie_2023}, expand the coding scheme with mood, action, scenery, etc., and quantify their correlations.
\subsection{A quantitizing approach to literary translation analysis}

This section describes two experiments on the applicability of MAQD for analyzing literary translations. 
The unit of analysis is a sentence pair in both, and the quantitization consists of distilling differences between versions of a text into either an open-ended description or a quantifiable categorical variable.
The data for the first task consists of the first paragraphs of G.~Orwell's "1984" (until the "war is peace, freedom is slavery"), the English original and its Italian translation sentence-tokenized and aligned using the LLM-powered 
BERTalign \parencite{liu_bertalign_2023}, yielding 47 pairs. GPT-4 was prompted to examine and highlight any major lexical or stylistic differences in each pair. The outcome was evaluated by two native Italian-speaking literature scholars (see Acknowledgments). Both concluded the alignment was correct and the inferences insightful, with no significant misinterpretations. 

The second task involves a simulated test set of 100 English-Japanese sentence pairs (about a rabbit in a forest, children's fiction genre, GPT-4-generated).
25 pairs match closely, but in 25 the rabbit is replaced with a bear in Japanese, in another 25 a moose is added, and in the last 25 the rabbit kisses his rabbit girlfriend, "redacted" in Japanese \parencite[emulating censorship scenarios not uncommon in the real world; cf.][]{inggs_censorship_2011}.
There are a few ways differences in text can be quantified. String edit distances are widely used in linguistics and NLP \parencite{manning_introduction_2008,wichmann_evaluating_2010}. Levenshtein distance is the sum of the optimal number of character additions, deletions or substitutions to transform one string (e.g. word) to another. These work within the same language, but cannot capture synonymy or cross-linguistic differences. Machine translation algorithms or multilingual sentence embeddings can output a numeric similarity between two sentences in different languages, but a more fine-grained, interpretable metric might be useful. 
The quantitization step consists of determining the value of such a "semantic edit distance" variable with four levels: no difference, addition, deletion, or substitution of contents or meaning. 
The dataset described above therefore has a ground truth total "distance" of 75/100. Further variables in real research could capture style, sentiment, author and translator metadata, etc., and associations between those could be statistically modeled.
GPT-4 results on the current task are very good: 96\% accuracy across the four classes (0.95 kappa; the simple sum of predicted difference classes is 74/100, just 1 off). 
Although small exercises, these examples show how combining multilingual LLM-powered auto-aligners with generative LLM-driven interpretation in a MAQD can enable scaling up the comparative analysis of translated, altered, or censored texts to larger text collections than a human researcher could manually process in a lifetime.

\subsection{Further case studies: automating interview analysis and stance detection, and analyzing idea reuse, semantic change, and visual data }

Given space limitations, the rest of the case studies are only briefly summarized here, with longer technical descriptions found in the SI.

\subsubsection{Machine-quantitized interview analysis, opinion and stance detection}

Interview-based,
phenomenological, and opinion studies across disciplines are often qualitative by label, yet can be observed at times giving in to the temptation of making quantitative claims about "often", "less", "many" etc. even when lacking 
systematic quantification or statistical modeling of the (un)certainty of such claims and the commonly repeated measures nature of the data \parencite[cf.][]{hrastinski_how_2012,norman_covid-19_2021,paasonen_about_2023}.
A more systematic, e.g. a QD approach would be preferable \parencite{banha_quantitizing_2022}. 
One such example is provided in the SI, illustrating how relevant passages (units) can be extracted from a larger text or interview and quantitized according to a coding scheme (e.g. a stance towards a phenomenon of interest, metadata variables about the respondents). This is followed by statistical modeling of the variables to test an example hypothesis that living arrangements are predictive of stance towards distant learning, controlling for participant age, and modeling the effect of measures (multiple expressions from each participant). The transformation of opinions into discrete variables can be outsourced to machines; in the example, GPT-4 detects the context-specific sentiment with virtually 100\% accuracy. As discussed above and in recent literature, this approach does not necessarily outperform bespoke supervised models, but is often easier to implement, as every variable to be coded does not require training or tuning a separate model, just good enough prompting of a capable pre-trained model.
If the source of data is a larger text corpus or semi-structured interviews as above, it is often necessary to first detect and unitize relevant passages.
In another upcoming paper, we will report on a cross-sector media monitoring project with the Estonian Police and Border Guard Board to analyze large text datasets for societal stances towards the police, combining keyword search and on-demand LLM-driven filtering and analysis. To briefly summarize here: 259 sentences collected from a newspaper corpus were annotated for evaluation, yielding 31 negative, 199 neutral, 19 positive, and 90 non-relevant sentences (e.g. containing police-related keywords but being about other countries or fiction or expressions like "fashion police"). In detecting relevance, GPT-3.5 had 76\% accuracy; the newer GPT-4 got 95\% (kappa=0.9, mean F1=0.9). For stance, GPT-3.5 matched human annotations 78\% while GPT-4 got 95\% (kappa=0.88; F1=0.92), reflecting results of similar benchmarking exercises \parencite[][]{zhang_sentiment_2024,gilardi_chatgpt_2023}, including in smaller languages \parencite{mets_automated_2024,rathje_gpt_2024}.

\subsubsection{Text and idea reuse analysis}

Political studies, history of ideas, cultural analytics, and science of science are among the disciplines interested in the spread and reuse of ideas and texts \parencite[see][]{chaturvedi_where_2018,linder_text_2020,salmi_reuse_2021,gienapp_large_2023}.
Automated reuse detection may be based on keywords, latent embeddings or hybrid approaches \parencite{chaturvedi_where_2018,manjavacas_feasibility_2019}.
While detecting verbatim reprints of an entire news article or passage is often easy, tracking the reuse and spread of smaller units and abstract ideas is hard, more so if it crosses language boundaries.
This exercise 
involved a simulated dataset of 100 social media post-like passages to test the feasibility of detecting the recurrence of a known pseudo-historical concept of "Russians are descendants of the Huns" \parencite{oiva_mapping_2022}.
The quantization step: evaluating if a passage (the unit of data) contains the concept. In a real research scenario, this could be followed by a quantification step to model e.g. the spread or change over time. GPT-4 is shown to complete the task with ease
(virtually 100\% accuracy), even after rephrasing, introducing OCR-like distortions, across translations between English and Russian --- and combinations of all of the above (see the SI for details and examples).

\subsubsection{Lexical semantic change detection}

Unsupervised lexical change or shift detection is a task and research area in computational linguistics that attempts to infer changes in the meanings of words over time, typically based on large diachronic text corpora \parencite{gulordava_distributional_2011,hamilton_diachronic_2016,dubossarsky_time-out_2019}.  
A SemEval 2020 shared task \parencite{schlechtweg_semeval-2020_2020} pitted numerous approaches against an annotated test of several languages and centuries of language change. The subtasks were: binary classification (which of the 27-38 test words per language have lost or gained senses between the given time periods?), 
and graded change detection (words ranked by change). Type-based embeddings were somewhat surprisingly still more successful in that task than contextual (BERT-like) models, which were later shown to require some task-specific engineering \parencite{rosin_temporal_2022}. The latter is the highest scoring approach (on task 2) on the same test set so far \parencite{montanelli_survey_2023}. 
A semantic similarity, divergence, or change analysis may be practical to formulate in MAQD terms. Here the SemEval task is replicated with GPT-4 as the annotator. The unit: a pair of sentences containing a target word from a given language, sourced from the paired corpora provided in the task. Coding: the DURel scheme used for the original ground truth data of the task \parencite[cf.][]{schlechtweg_diachronic_2018}. The quantitization: evaluate if the word occurs in the same or different meaning. Quantification: as per subtask, aggregation of change scores as binary decisions or a ranking. As further illustrated in the SI, this simple approach performs about as well as the best SemEval model in the binary task in English (70\% accuracy) and outperforms it in the ranking task ($\rho=0.81$ correlation vs 0.42, i.e. 2x improvement), as well as the more recent \textcite{rosin_temporal_2022} LLM-based result of 0.52. For German and Latin the results were not as good (see SI). This shows that a simple zero-shot machine-annotation approach can approach human accuracy in a difficult task like this, and can work on par or even surpass purpose-built complex architectures based on smaller LLMs or embeddings while requiring minimal effort to implement. This has also been further explored in recent works \parencite{periti_chatgpt_2024,ren_few-shot_2024,yadav_towards_2024} published between the preprinting and journal submission of the paper at hand.

\subsubsection{Zero-shot sense inference for novel words}

Another linguistic task that can benefit from QD thinking and machine automation is the semantic analysis of polysemous words, lexical innovations, or borrowings. The approach discussed above focuses contrasting potentially diverged or changed meanings. Another is the feature analysis in corpus linguistics discussed in the Introduction, where the componential meaning of (typically one or a group of) words may be analyzed via multiple fixed-level variables. The following is a more explorative task: inferring the meaning of previously unseen words. The test set of 360 sentences is again synthetic, split across three target senses (\textit{bear}, \textit{glue} or \textit{thief}; all represented by a placeholder nonce word), three languages (English, Turkish, Estonian), and two genres.
The evaluation was iterative for testing purposes: the annotator was shown an increasingly larger set of shuffled examples of a given sense.
The unit here is thus a set of 1-10 sentences, the coding scheme contains just the open-ended meaning variable, and GPT-4 was prompted to guess a one-word definition for the target placeholder given the context (without any further output constraints).
A subsequent quantification step could in principle quantify e.g. sense frequencies or trends over time or across genres and topics \parencite[cf.][]{karjus_quantifying_2020} to test linguistic hypotheses.
The annotation results, further illustrated in the SI, are promising: while \textit{bear} requires more examples (being confused with other large animals), GPT-4 can correctly infer the other two in all languages from just 3-4 examples. This shows both linguistics and applied lexicography can benefit from applying LLMs as machine assistants either in lieu or in conjunction with specialized models or human lexicographers \parencite{lew_chatgpt_2023,lew_dictionaries_2024}.

\subsubsection{Visual analytics at scale using multimodal AI}

\begin{figure}[htb]
	\noindent
	\includegraphics[width=\columnwidth]{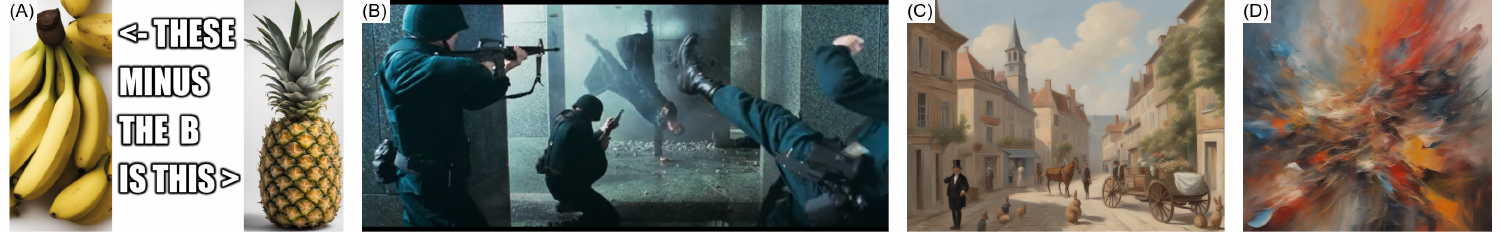}
	\caption{
    Examples of visual analytics using the vision-capable GPT-4o. \\
    (A) Prompt: Explain this joke in one sentence. --- Output: The joke plays on wordplay: removing the letter "B" from the word "bananas" leaves you with "ananas", which is another name for pineapple in many languages. \\
    (B) This is a still from a movie. Provide a list of likely genre, characters and their actions, location, mood, color palette, lighting, camera angle, shot type; all in variable: value format, short descriptions please. --- 
    Genre: Action, Sci-Fi
    Characters: Protagonist (dodging), Armed agents (attacking).
    Location: Urban, interior, possibly a government or corporate building.
    Mood: Tense, high-stakes.
    Color Palette: Cool tones, mainly blues and greens.
    Lighting: Low key, dramatic shadows.
    Camera Angle: Low angle, emphasizing action.
    Shot Type: Medium-wide shot, capturing multiple characters and action. \\
    (C) Make a nested list of the most prominent objects and subjects in this painting, output as single-line compact minified JSON, use short one-word or descriptive camelCase labels only. --- 
    ["streetScene", ["buildings", "horse", "cart", "people", ["manWithTopHat", "womanInWhiteDress"], "bunnies", ["brownBunny", "blackBunny", "whiteBunny"], "trees", "clouds", "churchWithSpire"]]
    \\
    (D) Describe the objective technical aspects of this painting like an art scholar, in one short concise paragraph. ---
    This painting features a dynamic composition of swirling, intermingling colors with a strong emphasis on movement and texture. The color palette is dominated by warm tones of orange, red, and yellow, contrasted by cooler shades of blue and grey. The brushwork appears loose and expressive, with layers of paint blending seamlessly into one another, creating a sense of depth and fluidity. The overall form is abstract, with no discernible figures or objects, focusing on the interaction between the colors and the implied motion within the composition.
	}\label{fig_visual}
\end{figure}

The case studies above have focused on the
capabilities of LLMs to assist with the annotation and analysis of textual data, a common medium in H\&SS. There is a clear direction towards multimodal models though, starting with vision. This includes the GPT-4 family of models \parencite{openai_gpt-4_2023}, several other commercial frontier models, and open models like CogVLM, BLIP and LLaVA,
which have all been shown to perform well in complex visual tasks \parencite{wang_cogvlm_2023,li_blip-2_2023,wu_gpt4vis_2023}.
Various studies have explored the potential for using such models in media and cultural topics \parencite{lyu_gpt-4vision_2023,smits_multimodal_2023,limberg_leveraging_2024,tang_leveraging_2024}. 
The MAQD is data-agnostic and suitable for conceptualizing the analytics of visual data, be it paintings, photographs, book covers, film scenes, or museal artefacts. As with text, it is important to determine the units and coding scheme, follow systematic quantitization procedures, and apply suitable statistical modeling to the inferred variables.
While additional visual case studies are not provided here, Figure \ref{fig_visual} depicts four examples of utilizing GPT-4o for image analytics. To avoid the possibility of the LLM drawing on any "memorized" training data, all images in Figure \ref{fig_visual} were generated (using Stable Diffusion XL1.0), except for The Matrix lobby scene still, captured by the author.
While these are all toy examples, scaling up such questions and inquiries to large datasets promises an unprecedented scale of analytics for fields like film studies, art history, visual anthropology, etc. 
Narrative descriptions of images may not be useful for quantification but illustrate the already available capacities of this class of models, which can reason about multilingual multimodal jokes (Figure \ref{fig_visual}.A) and produce coherent descriptions of realistic scenes as well as abstract visuals.
Detecting and segmenting objects on images \parencite[][]{chen_2d_2023,kirillov_segment_2023} or inferring art styles and aesthetics \parencite{mao_deepart_2017,karjus_compression_2023} is nothing new as such, but 
vision-language models enable qualitative "reasoning" (Figure \ref{fig_visual}.A,B) and zero-shot classification and annotation (\ref{fig_visual}.C). 
The example results are not necessarily perfect, but the usability of these early models as multimodal annotation assistants will likely improve.

\FloatBarrier
\section{Discussion}   

There is no substitute for expert human judgment, reasoning and calibration, when it comes to designing, conducting and interpreting research. However, human time is a bottleneck. With suitable instructable machines, expertise can be scaled up to enable research on much larger volumes of data, enabling researchers to elevate to more representative sample sizes as well as ask new questions that were perhaps not yet feasible in the past.
Humanities, social sciences, cultural analytics, and other neighbors of philosophy are very well positioned to leverage this opportunity, with long traditions in theory building, qualitative reasoning, and expert knowledge-driven analytics. 
The quantitizing design, exemplified by feature analysis in linguistics, provides a flexible and replicable framework.
These competencies are crucial for its application, which requires solid questions or hypotheses, a well-aligned coding scheme, expert-annotated test sets for evaluation, and meaningful interpretation of the results of quantitative modeling of the quantitized variables.
The MAQD is a scaling augmentation via automation and consequent error modeling. Here the machines of choice were instructable LLMs as flexible zero-shot classifiers and reasoners.

Continuing to use potentially quasi-quantifying designs would seem difficult to justify, 
when more systematic options are available, employing transparent procedures, replicable annotation and systematic quantification. Purely qualitative and conceptual research naturally has its place. However, applying qualitative designs in empirical scenarios where the actual (even if unstated) goal is quantification or extrapolation can lead to unintentional bad practices and spurious results. Where more data are available, limiting a study to tiny subsamples is no longer necessary, as previously time-consuming analysis tasks can be delegated to suitable machines. 
This all is not to say LLMs or ML or AI should be applied to everything everywhere all at once. Automating annotation is rather an optimization problem between available resources, human time, and output quality. However, as shown above, and in recent literature, using currently already available LLMs does not always decrease quality, and can in some cases even improve upon human assistance \parencite[][]{gilardi_chatgpt_2023,tornberg_chatgpt-4_2023,mellon_ais_2024}.

\subsection{Limitations and concerns}

There are naturally inherent limitations to using such technology. 
Technological ones include model performance and inhibiting guardrails. Current models are limited to 1-2 modalities, but e.g. 
natural human communication is highly multimodal \parencite[cf.][]{rasenberg_multimodal_2022}. This may well improve in the near future, however. One critique against using ML and LLMs to annotate data is that they can be unreliable or unreplicable and their outputs may be stochastic. Reasons include the nature of the underlying neural networks and loosely documented updates to cloud service LMMs. A related worry is that like all ML models, LLMs can inherit biases from their (in commercial cases, often unknown) training data \parencite{feng_pretraining_2023}. However, as discussed elsewhere \parencite{tornberg_chatgpt-4_2023,mellon_ais_2024}, these issues are not categorically unique to machines. They also apply to human analysts, crowd-worker annotators, research assistants, etc. Humans too are stochastic black boxes; engaging in analytic tasks requiring subjective judgments and reasoning can propagate and amplify biases. The solution in qualitative (and mixed, quantitizing) research is to be mindful, reflect and acknowledge this, and follow open science practices
(including preregistration) where possible to enable transparent scrutiny and replicability.

Using open-source or open-weight LLMs based on well-documented training procedures and data would be preferable in that regard \parencite[cf.][]{liesenfeld_opening_2023}. 
Running a fixed version of a local model can ease the replication issues that current cloud services may have if the model is public (in a persistent manner) or can be published along with the research. However, this is not always feasible \parencite{palmer_using_2024}, such as at the time of conducting the case studies here, where the only models capable of working with the smaller languages were the commercial closed-source ones.
One might also criticize using LLMs for the fact that usage has costs, either cloud service fees or investments into capable enough local hardware. The ecological footprint of using LLMs has also been raised. Then again, arguably any research activity has costs and a footprint \parencite[see also][]{tomlinson_carbon_2024}, including hiring research assistants or crowd-workers --- or using the most valuable resource, one's own time.

One way or another, LLMs are already being used in research, as discussed in the Introduction. The cost and effort of running a typical "paper-sized" study has significantly decreased in many disciplines, especially those not requiring experimentation or primary data collection, while LLM-based tools like ChatGPT or Copilot also expedite writing.
Anecdotally, the analytics for a typical feature-analytic linguistics paper involve annotating somewhere in the ballpark of 500 to 5000 examples, often sampled from a corpus; a PhD thesis about thrice that. Such a task can now be completed very fast using an LLM (at least at some level of quality).
A discipline or a journal that allows itself to be flooded by low-effort, low-insight papers is bound to eventually erode its reputation while hindering real progress.
Transparent practices and replicability have thus never been more important, and research evaluation might consider focusing less on volume and more on insight and intellectual contribution.

\subsection{Future research and opportunities}

While the case studies here aimed to cover a number of disciplines and task types, this contribution is by no means comprehensive in that regard.
Using LLMs and multimodal models as zero-shot classifiers and inference machines holds obvious potential for any field dealing with complex textual, visual, and otherwise qualitative data and products of human \parencite[and machine;][]{brinkmann_machine_2023} cultural evolution.
Already currently available LLMs can be plugged into research pipelines for classification, analysis, data processing and filtering. A single LLM prompt can often perform as well as complex multi-model pre-processing pipelines,
which were of course necessary up until very recently --- to the point of sometimes being research goals themselves \parencite[cf.][]{chaturvedi_where_2018,sherstinova_topic_2022,ash_relatio_2023,sobchuk_computational_2024}.
It may well make sense for researchers and groups to deploy custom model(s) on in-house hardware or private cloud, with fine-tuning for their domain and frequent use cases. It would be surprising if that would not become commonplace in the near future --- but with great data processing power comes quantitative and statistical responsibility.

There are various other domains where generative models may be useful. One is experiments employing artificial languages or visual stimuli, as used in psychology and cognitive science \parencite{kirby_cumulative_2008,galantucci_experimental_2012,tamariz_culture_2015,karjus_conceptual_2021}, or cases where examples need to be anonymized \parencite{asadchy_perceived_2024}, or open-ended responses analyzed. LLM-powered tools can be used to help build interfaces to run experiments. These are all tasks typically shared between a research team, but allocating some to machines means for example a cognitive scientist no longer needs to act as a full-stack developer, web designer, artist, and data annotator all in one. 

Other areas may include law, educational sciences or pedagogy. Empirical data like interviews, observations, practice reports but also laws and regulations could be systematically analyzed at scale in a MAQD. In an educational setting, LLMs may be useful for assessment and other tasks \parencite{baidoo-anu_education_2023,kasneci_chatgpt_2023}. 
LLM-driven chatbots can also be used to conduct semi-structured interviews \parencite{xiao_tell_2020}.
Another use case is potentially harmful, toxic, or triggering content: instead of subjecting a crowd-worker or RA to such a task, it can be delegated to a machine.

One framework that relies on (machine-assisted) quantification of textual data is distant reading \parencite{moretti_distant_2013}, typically culminating in interpreting word counts or topic distributions. Naturally, such representations are removed from the nuances of the content itself (the domain of "close reading").
One critique of Moretti-style distant reading \parencite{ascari_dangers_2014} states that its reliance on predefined genre labels and "abstract models that rest on old cultural prejudices is not the best way to come to grips with complexity."
In a MAQD, instead of 
relying on broad genre labels or abstract topic models, it is easy to model texts as distributions or sequences (of theory-driven units) at any or multiple levels of granularity, including where the volumes of text would be unfeasible for close reading; machine reading instead of distant reading.

\subsection{Time and efficiency gains}

\textcite{ziems_can_2023} suggest that the resources saved from allocating some tasks to LLMs would be put to good use by training expert human annotators or assistants. Indeed: let machines do repetitive labor and humans more interesting work. 
The time savings can be considerable. For example, the dataset of the first case study on newsreels features a modest 12707 synopses totaling about 281k words.
Assuming a reading speed of 184 wpm \parencite[words per minute; average for Russian language text;][]{trauzettel-klosinski_standardized_2012}, manual reading would be 19+ hours of work, with annotation likely taking as much again; easily one working week --- assuming the availability of a speaker of the language, 
able to interpret the historical context and the various abbreviations and references therein. Running it through an LLM service 
was a matter of leaving a script running for an hour or so --- yielding results very close to our expert with said qualifications.

The English translation of "Les Misérables" in the network example is about 558k words. Reading it would be 40 hours (English average of 228 wpm), and taking notes of all interacting character pairs would likely double that. Again easily two weeks of work. Or a few LLM hours or even minutes.
The Corpus of Historical American English \parencite[19-20th century;][]{davies_corpus_2010} is a common resource in linguistics. While NLP methods have been used to parse it for various ends, including the semantic change task discussed above, reading through its entire 400M words would take a human over 14 years (assuming 250 8h-workdays per year). No scholar in their right mind would undertake this, so either small samples or aggregation via NLP methods is used. Meaningfully analyzing every single sentence therein using an LLM is very feasible, without even needing to annotate training sets first.

The distant reading exercise purporting to launch a field of "culturomics" by analyzing "millions of books" \parencite{michel_quantitative_2011} was based not on "books" but rather aggregated n-gram counts. The dataset was large though: even just the English segment (361B words) would be 13,000 years worth of reading. 
Processing a book dataset of that volume in a MAQD would likely take more than a few hours, but is already feasible with current LLMs, and would enable asking more meaningful questions than word frequencies can provide.

\section{Conclusions}

Recent benchmarking studies have demonstrated the applicability of pretrained large language models to various data analytics and annotation tasks that until recently would have required human experts or complex computational pipelines. This contribution addressed the challenge of reliably making use of this automation and scaling potential in exploratory and confirmatory research, by advocating for a practical quantitizing-type research design framework. Its applicability in tandem with current LLMs as instructable machine assistants was assessed in an array of multilingual case studies in humanities, social science, and cultural analytics topics. While using machine (and human) assistants does pose some risks, they can be largely mitigated through expert oversight, theory-driven analysis and unitization principles, rigorous quantitative modeling with error rate incorporation to avoid overconfident predictions, and generally transparent procedures and open science practices.

\section*{Acknowledgments}

The author was supported by the CUDAN ERA Chair project for Cultural Data Analytics, funded through the European Union Horizon 2020 research and innovation program (Project No. 810961).
The author would like to thank Mila Oiva for collaboration on the newsreels topic classification example which became an extended case study in this contribution, providing expertise in interpreting the additional results and feedback; Christine Cuskley for the collaboration on the Twitter paper, one component of which was also used here as an expanded example; Priit Lätti for providing a version of the maritime wrecks dataset, Daniele Monticelli and Novella Tedesco for providing expert evaluation in the English-Italian translation task,  Laur Kanger and Peeter Tinits for discussions and for providing the test set for the historical media text filtering task,  Oleg Sobchuk and Artjoms Šeļa for providing a test set for the literary genre detection task, and Tanya Escudero for discussions that led to expanding the literature review. Thanks for useful discussions and manuscript feedback go to Vejune Zemaityte, Mikhail Tamm, Mark Mets and Karmo Kroos.

\printbibliography

 \newpage
 \FloatBarrier

  \addtocontents{toc}{\protect\setcounter{tocdepth}{1}}
 \part*{Supplementary information}

 \renewcommand{\thefigure}{S\arabic{figure}}
  \renewcommand{\thesection}{S\arabic{section}}
 \setcounter{figure}{0}
 \setcounter{section}{0}

\addto\captionsenglish{%
  \renewcommand{\contentsname}{}%
}

\vspace{0.5cm}

This Supplementary Information document contains 
further technical and methodological details on the case studies that did not fit in the Results section of the main text due to space limitations, and provides an overview table of the LLM evaluation results across all of the case studies. 
The second half extends the main Methods section, providing additional methodological details and suggestions for applying a QD or MAQD approach, including data preparation and unitizing, setting up LLM annotators, further statistics considerations, and ensuring transparency in research. 
Finally, a complete list of all the prompts used to instruct the LLMs in the case studies is provided at the end.

\section{Further case study details and task performance overview}

This section provides further details on the data, implementation, and results of the case studies, extending the Results section. Table \ref{table_scores} is an overview of the tasks and the accuracy scores of the employed machine annotators across all case studies.

\begin{table}[!ht]
  \caption{Summary of case studies in this contribution, emulating various humanities and social sciences research tasks. The asterisk* marks (partial) replications of other research. The Acc column displays raw accuracy of the best-performing LMM in the task, either quantitization or a preceding relevance filtering (compared to human-annotated ground truth; results marked with $\rho$ are Spearman's rho values instead of accuracy). The Adj column shows the kappa or baseline-chance adjusted agreement where applicable. The dash marks open-ended tasks.}
  \label{table_scores}
\setlength{\tabcolsep}{3pt}
\renewcommand{\arraystretch}{1}
\begin{tabularx}{\textwidth}{L{0.2\textwidth} L{0.12\textwidth} R{0.05\textwidth} R{0.05\textwidth} L{0.26\textwidth} L{0.23\textwidth} }
\hline 
Task & Language & Acc & Adj & Data domain & Complexities \\ 
  \hline 
  \multicolumn{6}{l}{Quantitization tasks (converting free text into fixed categorical or numerical variables) } \\
   \hline
  Topic prediction & Russian & 0.88 & 0.85 & Cultural history, media & Historical, abbreviations \\
  Event cause inference & Estonian & 0.88 & 0.83 & Maritime history & Historical, abbreviations \\
  Interview analytics & English & 1 & 1 & Discourse/content analysis & \\
  Usage feature analysis* & Eng (18\textsuperscript{th} c) & 0.94 & 0.89 & Linguistics, culture & Historical \\
  Text\&idea reuse & Eng, Rus & 1  & 1 & History of ideas & Multilingual \\
  Semantic variation & English & \textsuperscript{$\rho$}0.6 & & Sociolinguistics & Social media text, emoji \\
  Semantic change* & Latin, German, English & \textsuperscript{$\rho$}0.1\dots 0.81 & & Linguistics, NLP & Historical, contextual \\
  Novel sense inference & Eng, Est, Turkish & \textasciitilde1 & & Lexicography, linguistics & Minimal context \\
  Genre inference* & English & 0.8 & 0.73  & Literature & Books mix genres \\
  Translation analytics, censorship detection & Eng, Italian, Japanese & 0.96 & 0.95 & Translation studies, culture & Multilingual \\
  Social network inference* & English & * & * & Network science, literature & Many characters, ambig. references \\
  Stance detection & Estonian & 0.95 & 0.92 & Media analytics & \\
  \hline
  \multicolumn{6}{l}{
  Pre-quantitization tasks: relevant content detection, filtering, augmentation} \\
   \hline
  Relevance for stance & Estonian & 0.95 & 0.91 & Media analytics & \\
  Relevance filtering* & English & 0.92 & 0.82 & Text mining, history, media & Low quality OCR \\
  Data augmentation & Finnish & 0.72 &  & Media studies & Minimal context \\
  \hline
\end{tabularx}
\end{table}

\subsection{Stance and opinion detection}

In a recent paper \parencite{mets_automated_2024}, we investigated the feasibility of using pretrained LLMs for stance detection in socio-politically complex topics and lower-resource languages, on the example of stances towards immigration in Estonia. Estonian is not low-resource, in the sense that there are written corpora and some NLP tools available, but as the number of speakers is small (the population of Estonia is 1.3M), the amount of training data available is limited compared to English or German. We experimented with fine-tuning the previous generation of BERT-class models \parencite{devlin_bert_2019} on a hand-annotated dataset of thousands of examples of pro/against/neutral stances towards immigration.
The best-ranking fine-tuned RoBERTa model performed on par with a zero-shot GPT-3.5 approach (which had just come out, in late 2022). Naturally, zero-shot is a much cheaper alternative, obviating the need for costly manual training set construction and LLM fine-tuning (which either requires beyond consumer-grade hardware or paying for a cloud service). Emergent LLM applicability research reported similar results \parencite{zhang_how_2023,gilardi_chatgpt_2023}.

In another upcoming paper, we will report on a collaboration with the Estonian Police and Border Guard Board on a cross-sector project to analyze large media datasets to determine societal stances towards the institution of police and police personnel. Estonia is a multilingual society: while the media primarily caters to the Estonian-speaking majority, there are newspapers, TV and Radio stations in Russian as well as outlets with news translated into English. This necessitates a multilingual approach. We apply a pipeline similar to the immigration case study: a first pass of keyword search across corpora of interest followed by LLM-based filtering of the found examples for relevancy, and LLM-powered stance analysis applied to this filtered set. Finding contextually relevant examples from simpler keyword matches is crucial for accurate stance detection. For example, if the target of interest is Estonian Police, articles discussing police news from other countries, metaphorical expressions ("fashion police") and fictional contexts (films featuring police) should be excluded. 

While in the recent past accurate stance detection or aspect-based sentiment analysis would have required complex machine learning pipelines \parencite[][]{kucuk_stance_2020,nazir_issues_2022,rezapour_moving_2023} and model tuning, this can now be solved with zero-shot LLMs. We annotated a 259-sentence test set in Estonian; there were 90 non-relevant examples, and of the relevant 31 were negative, 199 neutral, 19 positive. This is quite representative, most police-related reporting is neutral about the police itself. In detecting relevant examples, GPT-3.5 only gets to 76\% accuracy (kappa=0.4, i.e. accounting for baseline chance; mean F1 at 0.54) but GPT-4 achieves 95\% accuracy (kappa=0.9, F1=0.9). We included both the target sentence and the title of the source article in the prompt for context, but some cases are difficult, e.g. where it is only implied indirectly that the police of another country is discussed. More context (e.g. paragraphs or a fixed larger content window) might help but is more costly (more tokens to parse). In stance detection, GPT-3.5 agrees with human annotations at a rate of 78\% (kappa=0.36; F1=0.51), while GPT-4 gets to 95\% accuracy (kappa=0.88; mean F1=0.92). Again, this is quite good, as many examples are ambiguous. For example, a sentence can be overtly negative while the police may be mentioned as a neutral participant; or the police might be reported to have done their job, which could be seen as neutral or positive, depending on perspective. Yet LLMs show promise as universal NLP tools in media monitoring contexts.

\subsection{LLMs for missing data augmentation: Finnish broadcast data example}

This task is not an example of MAQD but rather a preprocessing step that is often necessary to carry out a multifactorial analysis when the data is sparse or otherwise incomplete; as such it is left to the Supplementary. In a recent public service media study \parencite{ibrus_quantifying_2022}, we explored a large dataset of television metadata from a broadcast management system (BMS; essentially a production database as well as an archive) of the channels of the Estonian Public Broadcasting (ERR), covering 201k screen time hours across 408k program entries between 2004-2020. In a follow-up study (in prep.), this is being compared to a similar BMS dataset of neighboring Finland's broadcaster YLE. Such data are often sparse, with notably production countries missing in this one in about 23\% of daily programming entries. The missing data could be manually inferred or searched for using the available program title, short synopsis, and secondary sources, but this would of course be incredibly time-consuming. This prompted an experiment to infer the production country directly from these limited metadata using GPT-4, using a randomized test set of 200 unique program entries where the true production country was actually present (15 different countries ended up in the test set). 

Unlike in most other classification tasks here, the missing data imputation task was set up without constraining output classes to a fixed set, to give the LLM free reign to take an (artificial) educated guess. The task is further complicated by the fact that the entire BMS is in Finnish language, including the titles and the (very) short descriptions that are used to prompt the LLM. 
Besides many country names like \textit{Yhdysvallat} (USA), smaller place names can also be translated, e.g. Lake Skadar (also Skadarsko, Scutari; in the Balkans) is referred to as \textit{Skutarijärvi} in one of the synopses, which could as well be a Finnish place name.
Non-Finnish names are also modified according to Finnish morphology, e.g. \textit{Jamie odottaa tuomionsa toimeenpanoa \textbf{Wentworth}in linnassa, mutta pian häntä odottaa kuolemaakin kurjempi kohtalo. Claire panee henkensä likoon pelastaakseen miehensä sadistisen \textbf{Randall}in kynsistä.} This synopsis is also typical in length; the median in the test set is 206 characters.

Despite these complexities and the open-ended nature of the task, the results are promising, with GPT-4 yielding 72\% accuracy. Most mismatches make sense too: mixing up English-speaking countries is the most common source of errors, followed by the German-speaking, and the Nordic countries. This illustrates the applicability of LLMs for data imputation and augmentation in complex social and media science datasets, but also the necessity to account for error rates in any subsequent statistical modeling based on the augmented data, to avoid biases, as discussed in Methods. If augmented data are added to an existing database, it should of course be transparently flagged as such.

\subsection{Text and idea reuse detection}

A synthetic test set of pseudohistory-style blog posts is used here, modeled directly after \textcite{oiva_mapping_2022}, who surveyed the landscape of pseudo-historical ideas and their outlets in Russian-language online spaces.
The classification tasks involve detecting the occurrence or "reuse" of the idea that "Russians are descendants of the Huns", apparently common in such user groups.
The data is generated as follows. GPT-4 was first instructed to compile 50 short paragraphs in English on various other pseudohistorical topics drawn from \textcite{oiva_mapping_2022} that would include this claim and 50 that would not. These 100 items were then modulated and distorted in a variety of ways, again using GPT-4: rephrasing the claim, inducing "OCR errors", translating into Russian --- and combinations thereof. As an example of original text and its maximal modulation: \\
\textit{It's an often-overlooked fact that all the weapons used in seventeenth-century Europe were produced by the Russians. This massive weapons production and export reflect an advanced civilization, attesting to the fact that Russians are descendants of the Huns. It's a narrative that resists the distortions of history, reaffirming Russian heritage.} This becomes:\\
\fontencoding{T2A}\selectfont
\textit{Эт\textup{o} част\textup{o} пр\textup{e}н\textup{e}бр\textup{e}г\textup{a}емый ф\textup{a}кт, чт\textup{o} вс\textup{e} \textup{o}ру жи\textup{e}, использованное в \textup{E}вр\textup{o}п\textup{e} \textup{XVII} в\textup{e}к\textup{a}, был\textup{o} пр\textup{o}изв\textup{e}д\textup{e}н\textup{o} русскими. Эт\textup{o} м\textup{a}сс\textup{o}в\textup{o}е пр\textup{o} изв\textup{o}дств\textup{o} и \textup{e}ксп\textup{o}рт \textup{o}ружия свид\textup{e}т\textup{e}льствуют \textup{o} развит\textup{o}й цивилизации и подтв\textup{e}ржд\textup{a}ют, чт\textup{o} кровь гунн\textup{o}в теч\textup{e}т в в\textup{e}н\textup{a}х русских. Эт\textup{o} пов\textup{e}ств\textup{o}в\textup{a}ни\textup{e} с\textup{o}пр\textup{o}тивл\textup{y}ется искажениям истории, подтв\textup{e}ржд\textup{a} русск\textup{o}е насл\textup{e}ди\textup{e}.}
\fontencoding{T1}\selectfont
The "OCR distortions" may not be immediately noticeable, consisting mostly of swapping out Cyrillic letters with similar-looking Latin ones and introducing spaces (both of which would easily confuse simpler e.g. keyword or string distance-driven classifiers).

The results are easy to report: GPT-4, with instructions to assess if the claim is present, had a perfect 100/100 accuracy rate in most conditions. There was only one case where it failed, where the initial generation prompt had introduced a negation: \textit{/\dots/undermines the Russian heritage by insinuating that Russians are not descendants of the Huns}. It was not an issue in test cases that had undergone translation to Russian. A detection prompt could of course be better crafted to also catch such items. In summary, as shown here, text and idea reuse detection is very much feasible using instructable LLMs such as GPT-4, including cases of idea transfer across languages.

\FloatBarrier

\subsection{Example LLM-powered interview analysis}
 
The data in this exercise are synthetic, generated using GPT-4, which was promoted to output a variety of responses, as if uttered by college students, concerning the benefits and downsides of doing group assignments online as opposed to meeting live. This emulates a scenario where the researcher has already unitized and filtered the interview data as relevant passages or responses. The latter step could be done either manually, by searching for keywords or by using another machine classification. 
The synthetic data includes examples such as: \\ 
\textit{You know, one of the things that bothers me about online meetings is that it's harder to have those spontaneous moments of laughter or fun that make the work enjoyable, and that's something I really miss.} \\
The data of 192 passages was simulated so that there are multiple responses from each "student" (n=56, random ages), some living on-campus and some off, with a simulated difference in stances. In the resulting data, the 109 off-campus student responses are split 36/73 negative-positive; 64/19 for on-campus. 
Given the simulated nature of the data, the main variable of interest, stance towards online group work, is already known (as if coded by a human annotator).

The (admittedly simplistic) example hypothesis is: controlling for age, students living on campus see more negative aspects in doing group assignments online than those off campus. This can be tested using a mixed effects binomial regression model; the random effects structure is used to take into account the repeated measures. The model can be conveniently run using the popular lme4 package in R with the following syntax:
\begin{align*}
&\text{Logistic regression:} && \log\left(\frac{p_{ij}}{1 - p_{ij}}\right) = \beta_0 + \beta_1 \cdot \text{campus}_{ij} + \beta_2 \cdot \text{age}_{ij} + u_{j} \\
&\text{In lme4 Syntax:} && \text{online} \sim \text{campus} + \text{age} + (1|\text{id})
\end{align*}

In the constructed model, "off-campus" is the reference category for the campus variable and "off" for the response. Living on campus is associated with a decrease in the log-odds of a positive response $\beta=-1.9$ or $e^{1.9} = 0.14$ times, $p<0.001$. Regression model assumptions were checked and were found to be met. The p-value indicates the probability of observing that effect is exceedingly small (0.0000000438) if the null hypothesis (no effect of living on campus) was true, so the alternative hypothesis (on-campus students don't like online) can be accepted.

This test was conducted directly on the synthetic data, equivalent to a scenario where a human annotated the interpretations. The same could be completed by an LMM instructed to determine the stance or attitude of the student towards online group assignments in each response (regardless of the overall sentiment of the response). The LLM accuracy results are easy to report here: a suitably instructed GPT-4 detected stance towards online learning from the narrative-form responses with a 100\% accuracy; i.e. the machine interpretations did not differ from ground truth in this case. Note that this exercise was independent of the data synthesis. The fact that GPT-4 both generated the initial data and was used for classification has no bearing on the accuracy, which likely stems from the combination of relatively clear stance expressions in the data and the ease of inferring them for GPT-4.
If the accuracy was considerably lower --- in a real research scenario, measured using e.g. a small human-annotated test set --- then the error rate should be incorporated in the regression modeling to avoid biased estimates. See the Methods section for one approach to how to do that.

While this example focused on a confirmatory case, interview-based research could benefit from machine assistance on other tasks as well. The retrieval of relevant examples was mentioned; another may be clustering interviewees \parencite[cf.][]{kandel_enterprise_2012}. This could indeed be done using e.g. latent topic models, but as in the confirmatory topic classification example, can be approached in a more principled way by having the LLM annotate responses for specific themes of interest, and then using those for clustering.

\subsection{Lexical semantic change detection}

\begin{figure}[htb]
	\noindent
	\includegraphics[width=\columnwidth]{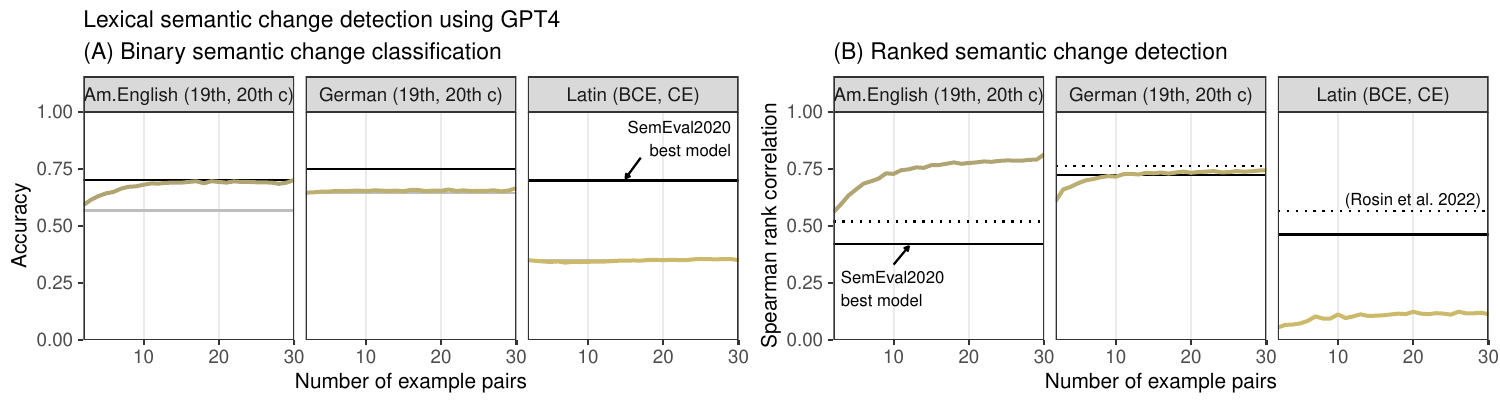}
	\caption{Lexical semantic change detection using GPT-4 on two tasks, binary classification (A) and graded change (B), in three languages. The trend lines illustrate how well the zero-shot approach performs given an increasing number of example pairs (bootstrapped average values).
    The top results from the SemEval task are highlighted with solid lines.
    The gray lines are random baselines for binary classification.
    A later LLM-based result in (B) is shown with the dotted line.
    GPT-4 performs best in English as expected, even surpassing past approaches on the second task.
	}\label{fig_semchange}
\end{figure}

This section provides more details on the results of the semantic change detection exercise.
In the binary classification task, the zero-shot approach, based on evaluating just a handful of examples, performed as well as the state of the art best model reported in the SemEval task in English (70\% accuracy; Figure \ref{fig_semchange}.A). 
It was just above the random (majority) baseline but below SOTA in German; and practically at random for Latin.
For this task, a threshold of 2 or more DURel-coded "unrelated" judgments was used (optimized thresholds could improve results). 
In the second, semantic change subtask however, it went well beyond the best SemEval result for English ($\rho=0.81$ vs 0.42, a 2x improvement; Figure \ref{fig_semchange}.B). It also surpassed the more recent \textcite{rosin_temporal_2022} LLM-based architecture that had a 0.52 correlation with the test set.
In German, the result of 0.75 is between that and the best SemEval model (0.76 and 0.73, respectively). Judging by the trend, it may improve if given more than just 30 examples (the SemEval models used entire training corpora; better prompting may also increase scores). Latin at 0.1 performed below both comparisons but above random.
For further context, other specialized LLM fine-tuning architectures have gotten as high as 0.74 on the same task in a different Indo-European language, Spanish  \parencite{zamora-reina_lscdiscovery_2022}.

Setting up this experiment here required writing a one-sentence prompt to be iterated with the example pairs on the GPT-4 API. In contrast, the authors of the models featured in the SemEval task paper \parencite{schlechtweg_semeval-2020_2020} clearly put no small amount of work into developing their various embedding, ensemble and LLM based architectures (each spawning at least one paper on their own). \textcite{rosin_temporal_2022} is a full-length paper in a high-ranking NLP conference.
On the flip side, pretrained instructable LLMs are only as good as their training data. Clearly, there was enough Latin in the GPT-4 training data for it to perform well here.

\subsection{Challenging linguistic data annotation}

In a separate study focusing on linguistic divergence in US American English \parencite{karjus_evolving_2023}, we looked into modeling differences between two groups of users, those aligned with the political "left" and those with the "right". 
The data was mined from the social media platform Twitter (now "X").
We experimented with using word embeddings of the type that performed well in the shared task discussed above \parencite{schlechtweg_semeval-2020_2020}, as well as annotating a small dataset by hand following the DURel framework \parencite[cf.][]{schlechtweg_diachronic_2018} mentioned above for evaluation purposes.
This involved comparing the usage and therefore meaning of a target word, phrase or emoji in contexts derived from the tweet corpus, for example (the examples have been rephrased here in order to preserve author anonymity): \\
\textit{We have a kitten right now who is in a bad condition, need to get him to a \textbf{vet}, got many more here like this} --- compared to ---
\textit{This is a president that knows how to withdraw forces when necessary. Perhaps if more \textbf{vets} ran for office we would have people in charge who can do what is needed}.

We also applied GPT-4 to the same annotation task in the role of a "third annotator". There were 8 target words and emoji, three comparison sets for each to determine difference as well as in-group polysemy; 320 pairs in total. The two human annotators had a good agreement rate of $\rho=0.87$ (measured in Spearman's rho, given the ordinal DURel scale). GPT-4 achieved moderate agreement of $\rho=0.45$ with one and $0.6$ with the second annotator. 
The lower rate compared to the human inter-rater agreement was partially affected by the emoji in the test set, which the humans also found difficult to annotate. There was also very little context to go on, and social media texts can include rather non-standard language and abbreviations. 
This exercise did not involve iteratively improving the simple single-sentence prompt --- better agreement may be achieved with more detailed, and potentially iterative or step-wise prompting.

\subsection{Zero-shot sense inference for novel words}

The synthetic test set for this exercise was generated also with GPT-4, instructed to compile a set of unrelated sentences that would use one of these three target senses: \textit{bear}, \textit{glue} and \textit{thief} (representing both animate and inanimate subjects, countable and mass nouns), in three languages: English, Turkish and Estonian (representing different language families and speaker population sizes). 
Each target sense is instructed to be expressed with a placeholder word, \textit{zoorplick}, instead. This was chosen as a word unlikely to exist in the training data. GPT-4 was also queried to guess its "meaning" and the machine came up with nothing. The context is intentionally just a sentence to make the task harder.

\begin{figure}[htb]
	\noindent
	\includegraphics[width=\columnwidth]{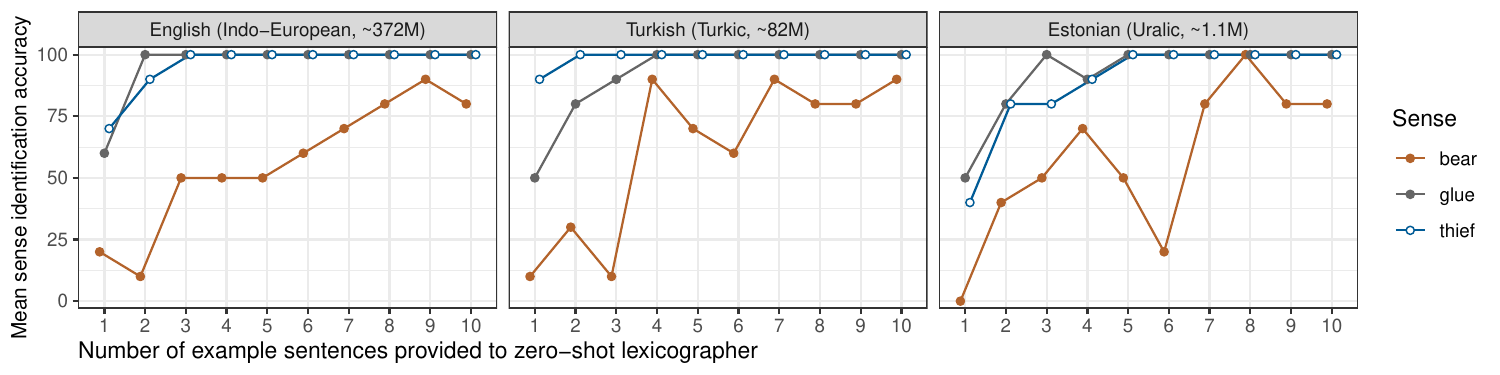}
	\caption{Three senses of a novel "word" inferred in three languages using GPT-4. In most cases, about 3-5 example sentences is enough for the LLM to correctly infer the intended novel meaning of the placeholder nonce word.
	}\label{fig_lexico}
\end{figure}

In the testing phase, GPT-4 was instructed to infer the meaning of the placeholder given the separately generated contexts. 
The LLM output was not constrained, making this an open-ended exercise
Some leeway was given: \textit{adhesive} would be accepted as correct for \textit{glue}, and \textit{burglar, robber, pickpocket} as types of \textit{thief}.
The results, illustrated in Figure \ref{fig_lexico}, are promising: \textit{glue} and \textit{thief} can be correctly inferred in all three languages already based on 3-4 examples. \textit{bear} is more difficult, as with only sentence-length contexts, the LLM mistakes it for various other wild predators, but accuracy improves with more examples.

\section{Additional methodological consideration for applying a MAQD}

This section extends the Methods section of the main text, starting with the method comparison, which, while clearly a simplification, may help position and situate the MAQD in relation to other designs commonly used in H\&SS.
\begin{itemize}[topsep=0pt,itemsep=0pt]
\item Qualitative designs:
    \begin{itemize}[topsep=0pt,itemsep=0pt]
    \pro Typically deeply focused, can consider wider context, reception, societal implications, etc. and self-reflections by the author.
    \con Hard to generalize and estimate the uncertainty of claims; typically hard to replicate, practically impossible to reproduce; involves inherently subjective analysis.
    \con Very hard to scale to large data.
    \end{itemize}
\item Quasi-quantifying designs:
    \begin{itemize}[topsep=0pt,partopsep=0pt,itemsep=0pt]
    \pro Same as above; focused, contextual, reflexive, etc.
    \pro Systematic coding schemes, if present, make it easier to replicate (if documented), but relationships and their uncertainty remain impressionistic.
    \con Otherwise same downsides as above, and hard to scale. Overconfidence in quantitative results without uncertainty modeling can lead to spurious results.
    \end{itemize}
\item Primarily quantitative methods:
    \begin{itemize}[topsep=0pt,partopsep=0pt,itemsep=0pt]
    \pro Applicable to big data and scalable; relationships and their uncertainty can be estimated; may be seen as more objective.
    \pro Easier to replicate (or reproduce if data and procedures are all made available).
    \con May lack the nuance and depth of qualitative analysis of meaning, context, and power relationships, especially for complex societal or cultural phenomena. Only applicable to counted or directly quantifiable data types.
    \end{itemize}
\item Quantitizing (qualitative to quantitative) designs, e.g. feature analysis:
    \begin{itemize}[topsep=0pt,partopsep=0pt,itemsep=0pt]
    \pro Inclusion of the qualitative step comes with most if not all benefits of qualitative-only analysis; including the ability to handle virtually any human-readable data type.
    \pro While the qualitative step involves subjectivity, it can be replicated, and the quantitative reproduced (given data and procedures); relationships and their uncertainty can be estimated.
    \con Hard to scale to large data.
    \end{itemize}
\item Machine-assisted quantitizing designs:
\begin{itemize}[topsep=0pt,partopsep=0pt,itemsep=0pt]
    \pro All benefits of qualitative analysis.
    \pro All benefits of QD, systematic quantification, applicability.
    \pro Yet applicable to big data and scalable.
    \end{itemize}
\end{itemize}

Applying the MAQD requires well-thought-out data sampling and unitization, a coding scheme with relevant variables, principled qualitative analysis or annotation procedures, and systematic quantitative (statistical) modeling of the quantitized variables. Additional aspects of these steps, extending the Method section, are covered below.

\subsection{Data preparation and unitization} 

Unitizing is a crucial step in data without what could be seen as natural units \parencite[for good practices, see the references in the Introduction and][]{krippendorff_content_2019}. It can be helpful to think of units as rows in a table where the columns are the variables and the first column contains the example units. Given an art collection, a painting is likely a useful unit of analysis. There may be multiple paintings per artist, but the unit is fairly non-controversial, and the subsequent statistical analysis, even if the goal is to compare said artists, can and should take into account this grouping of units (see mixed effects modeling discussion below). In contrast, an entire book can but is unlikely to be a useful unit that can be distilled into a single data point in a variable --- unless the goal is just to count pages, authors, or other variables applying to an entire book (but finer unitizing may well lead to better results in the latter as well). If the interest is in content, a likely unit of comparison is a paragraph or a sentence. The same applies to interview-based research: the unit, the data point, is unlikely to be an interview or a respondent, but all their (topic-relevant) utterances or answers (which can be grouped by respondent in the statistical modeling step).

\subsection{The coding or quantitization scheme} 

A coding scheme consists of variables and their definitions (see literature cited in the intro). Each categorical variable has a set of predefined levels (values) and their definitions. The scheme may be entirely or partially derived from preceding research and theory, or engineered by the domain expert from scratch for a given study. The qualitative analysis proceeds according to this scheme, but the scheme may be and in practice often is iteratively improved based on small initial data samples or a pilot study \parencite[][]{schreier_qualitative_2012}. The number of levels of a categorical variable are fixed and typically kept to a minimum, to ease interpretation of the quantification step.

For example, if the data are newspaper texts, the unit a paragraph, and the hypothesis that negative stances are foregrounded, then the variables and levels might be the dependent variable of stance (positive, negative), the main predictor the page (numeric; or binomial, front page or not), perhaps a control variable for type (news, opinion), and variables for text author and publication. The first three would be considered fixed effects and the last two random effects in the mixed effects statistical modeling sense; these would need to be ideally controlled for in the presence of repeated measures (which is more often than not the case in H\&SS research; see below).

For an overview on how to set up a zero-shot annotator machine, and the importance of a good coding scheme, see the relevant section in the SI.

\subsection{Setting up a zero-shot annotator}

While any suitable machine learning or statistical machine can be plugged into the MAQD framework, this section focuses on instructable LLMs in a zero-shot learning context as the currently most flexible option, forgoing the need for annotating training data for each new variable. The case studies are not focused on model comparison \parencite[like e.g.][]{ziems_can_2023,bandarkar_belebele_2023}. 
Two models were used here, primarily OpenAI's GPT-4, with occasional comparisons with the previous-generation GPT-3.5.
The model choice is mostly for practical reasons. Running inference on this cloud service is easy and fairly affordable, and does not require setting up a local LLM, which would require either hardware beyond the consumer grade, or a suitably powered and configurable cloud service. 
The GPT-4 class models are also highly multi-lingual, compared to most current open-source alternatives, which, based on limited attempts, still did not recognize the smaller languages of the intended case studies. 
However, more and larger open-source models are continuously becoming available,
and optimization research is ongoing to make them run on resource-constrained hardware \parencite[e.g.][]{taori_stanford_2023}. For simpler tasks in large languages like English, using smaller models (on the order of 7-8B parameters) is already feasible, and such models can run even on consumer-grade hardware or within the free allowance of services like Google Colab.
This section uses the cloud-based GPTs as an easy example, but the suggestions should be fairly generalizable.

In the case of the OpenAI (and similar) models as a service, analyzing texts consists of making a call to an API endpoint, which consists of a number of (fairly well documented) parameters. The associated Python packages \textsc{openai} and \textsc{tiktoken} can be freely used for easy implementation, which also makes it easy to keep an eye on costs (which are calculated per input and output tokens). 
While prompts can be tried out over the web-based ChatGPT interface, a chatbot is obviously not well suited for systematic data analysis, and likely has an above-zero temperature setting (its "code interpreter" plugin, later relabeled as just "data analysis" and finally integrated into the chatbot, was not found suitable either).
Currently, some programming is required to use these models, both cloud and locally-run LLMs, but this is becoming less of an issue as this technology is gradually integrated into various services and software (like MAXQDA, NVivo, and spreadsheet apps). This contribution also comes with an open code base to foster replication and enable easy experimentation with these tools.

The simplest input prompt for a suitably instruction-tuned model such as GPT-4 consists of just the instructions and the data to be analyzed, for example, \textit{Tag this text as positive or negative sentiment. Text: I love ice cream.} Other models \parencite[e.g.][]{peng_rwkv_2023} may require specific formatting to elicit instruction-following behavior, e.g. ending the prompt with \textit{\dots Assistant:} to be then completed with the response.
For more specific questions and concepts, it is useful to define the meaning of the output classes, rather than leave them for the language model to decipher. Using common language synonyms with definitions instead of precise but domain-specific terminology also worked better in some cases.
Multiple examples can be concatenated into a single prompt with a request to output multiple tags, but this can easily degrade classification accuracy and induce hallucinations --- these are, after all, just text generation engines. This appeared less of a problem for GPT-4 than 3.5, and may be worth experimenting with as a cost-optimization strategy. Some cloud LLM services are now also offering asynchronous "batch APIs", where a larger number of requests can be sent as a package and results retrieved at a later time with discounted costs. The prompt caching feature is also becoming more widespread and can save costs in an analytics scenario where the coding instructions are only "billed" once.
If the input data is long, e.g. an entire book, then the window size of the chosen model must be kept in mind (this is less of a worry with frontier models that have context windows of tens if not hundreds of thousands of words). Inputs that do not fit into a single prompt can be chunked and their result later aggregated. In most practical applications however, proper unitizing (e.g. paragraphs or chapters instead of entire books) is expected to yield more useful and fine-grained results anyhow.

Relevant parameters for a classification task include temperature, bias and output length \parencite[for details, see this white paper:][]{openai_gpt-4_2023}. These may slightly differ between models but the same principles hold for most current generation LLMs. 
"Temperature" controls the randomness or variety of the output of a neural network: in a chatbot like ChatGPT, a moderate to high value is desirable, while 0 usually makes sense for classification. Defining token bias allows for steering the model towards generating certain outputs with a higher probability. This is useful in a classification scenario where the output should be one of the class labels (setting their token values to 100 worked well), but should not be used where an open-ended output is expected. 
Finally, it is useful to limit model output to the maximum (token) length among the class labels, to make sure the model, generative as it is, does not add commentary and that the output is easy to parse (using single-token class labels where possible worked well). If using prompting strategies like chain-of-thought etc. \parencite{zhang_investigating_2023,chen_when_2023}, longer outputs must be allowed of course, and parsed accordingly. One option is to instruct to output a machine-readable format such as JSON \parencite[for a guide, see][]{ziems_can_2023}.
In this study, fairly short and simple, often single-sentence prompts were used (all prompts are at the end of the SI; short inputs also save on cloud service usage costs). 

One way or another, if a generative LLM is used as a classifier or annotator, it is important to keep in mind that it is actually not a classifier in the traditional ML sense, and may generate nonstandard outputs. This issue may also arise when the LLM or the service hosting the LLM detects potentially sensitive or harmful content in the input and refuses to respond. The extent of this varies, but it can also hinder using the model as a classifier in valid contexts: if a given input with potentially sensitive content triggers such an adverse reaction, the model may refuse to respond or respond with a refusal.
In any case, it is good practice to build contingencies for that into the pipeline.

This is the technical side of things. The most important component however --- in any quantitizing mixed designs including usage feature analysis --- is the qualitative coding scheme design, which precedes any annotation work. In the MAQD case, this also involves translating the coding instructions into a good prompt. In turn, the prerequisite for a good scheme and variables is a good question or hypothesis. The machine-assisted step can only scale up the qualitative expertise, and the quantification step can only estimate uncertainty to make sure the claims are reasonably likely to replicate (see next section). LLM tech at its current stage is unlikely to be a substitute for this careful and systematic qualitative work, theory grounding, and expert knowledge, that precedes and follows the fairly straightforward data annotation process and statistical machinery in the middle.

\subsection{Necessity for statistical modeling in empirical research}

\begin{figure}[htb]
	\noindent
	\includegraphics[width=\columnwidth]{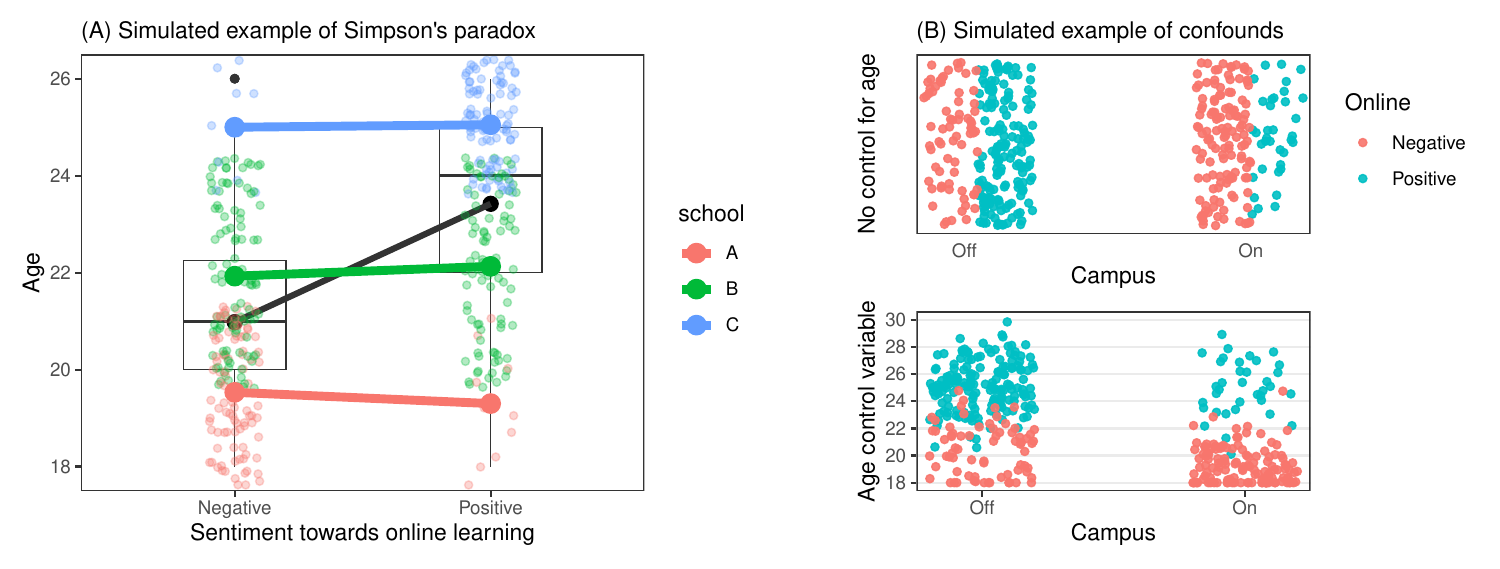}
	\caption{Illustrations of Simpson's Paradox (A) and the lack of control variable problem (B).
	}\label{fig_sup_simpson}
\end{figure}

As discussed in the Methods section in the main text, statistical modeling is a necessary step in any design dealing with quantitative or quantitized data. The following expansion of the Method section is relevant for fields not routinely engaged in systematic statistical modeling, but likely not interesting for practitioners who already are.
Figure \ref{fig_sup_simpson} illustrates two scenarios where making impressionistic conclusions or even limited statistical modeling can lead to spurious results. The data are permutations of the same synthetic interview data as used in the interview analysis case study. Here, the same data generation process has been manipulated to create two new exemplary scenarios. Figure \ref{fig_sup_simpson}.A illustrates the Simpson's paradox. The data is arranged as if collected from three schools or colleges. For simplicity, here it is assumed every student is asked only once (which would necessitate a more complex, nested random effects design). If the school grouping is ignored, it would be easy to conclude that younger students are more negative and the older more positive about online learning (black line, dots are averages). A logistic regression model (age predicting sentiment) would confirm this with a $p<0.0001$. However, when the random effect of school is added, in a mixed effects framework, the effect of age disappears ($p=0.28$). The graph illustrates why: the primary difference is between schools (some of which happen to have students of different ages and differences in sentiment toward online).

Figure \ref{fig_sup_simpson}.B illustrates how not controlling for a relevant confound can lead to spurious results. Here the data is generated without repeated measures, with a different age distribution, and with an added variable of campus (on or off-campus students). The top graph depicts the data without age (points are distributed randomly across the y-axis). This looks like more off-campus students are positive about online learning, while on-campus students are mostly negative. Again, a fixed-effects-only logistic regression would even confirm this (campus effect $p<0.0001$). However, adding age to the model --- depicted as the y-axis on the bottom plot --- clarifies the picture: it is first and foremost older students who like online learning (age $p<0.0001$, campus $p=0.426$). They are also indeed more likely to live off-campus, but it would not be possible to reject the null hypothesis (no campus effect on sentiment) based on these data.

\subsection{Ensuring transparency and replicability of qualitative and machine-assisted research}

One criticism that can be raised against qualitative research, quantitizing designs, as well as any machine-assisted designs, is that they are not completely transparent and replicable. All of these approaches involve subjective decision-making, and therefore annotator and analyst biases, and inherent stochasticity in the case of machine annotators, if models that are not 100\% deterministic are used.

The way to get around these issues and still allow for research on cultural, humanistic, and qualitative data, is to strive towards maximal transparency and good open science practices in all phases of a given study \parencite{mckiernan_how_2016,vicente-saez_open_2018,nosek_preregistration_2018,kapoor_reforms_2023}. In the QD case, this includes describing and publishing the coding scheme and unitization principles (possibly as a pre-registration), the annotated or analyzed unitized data, and code or instructions to reproduce the quantitative analysis step. MAQD adds the need to publish the prompts and information about the model that was used. As open-source LLMs become more capable and available, it is not unfeasible that the open data accompanying a study would include the (potentially fine-tuned) model variant as well.

In cases where the source data itself is of a sensitive nature and cannot be publicized, the coded variables (which typically do not reveal identities in the source) are still enough for the quantification and subsequent interpretations to be reproducible. In cases where even that would be an issue, synthetic or anonymization methods can be used to generate comparable data \parencite{james_synthetic_2021}.

To avoid underestimating the model error rates and subsequent uncertainty estimates (discussed above), setting up the test dataset can follow the proven machine learning philosophy of using independent training, evaluation, and test sets. In the zero-shot case, there is no training set, but any prompt engineering should be ideally completed on a separate evaluation set, only then to be tested (once) on the test set, to avoid overfitting and biasing confidence of the eventual analysis where the test set results will be used to estimate confidence or uncertainty. For more benchmarking considerations, see \textcite{hagendorff_machine_2024}.

There are large discrepancies between disciplines when it comes to open science practices. While publishing data and the concept of replicability are still practically unheard of in some, others like psychology have learned their lessons from the "replication crisis" \parencite{shrout_psychology_2018,scheel_excess_2021}. It is vital to adopt transparent practices when using the MAQD to avoid such pitfalls.

\subsection{Proficiency-based limitations in applying a machine-assisted framework}

The SAGE Handbook of Mixed Methods in Social \& Behavioral Research \parencite{tashakkori_sage_2010} lists the following hindrances to mixed methods research: "costs of conducting it, unrealistic expectations regarding an individual researcher's competence in both QUAL and QUAN methodology, complexity of putting together teams to carry out such research when groups (rather than individuals) conduct MMR, and (last, but not least) the impossibility of an individual or even a team's examining issues from different perspectives/worldviews."
The same, by extension, applies to the MAQD framework. While zero-code applications may well become available in the future, the low-code pipeline described in the Methods does require some proficiency in a suitable programming language, and in either using APIs or deploying local models. The quantification step furthermore necessitates a basic understanding of statistics and the skills to conduct modeling in a software package or a programming language like R or Python. There are two options here: either the scholar takes time to learn the basics of programming and statistics, or, collaborates with somebody who already has them. 
However, investment in learning (and teaching students) basic programming and statistics is worthwhile, with the added effect of broadening career prospects.

\newpage

\section{List of prompts of all case studies in this paper}

This section lists all the prompts that were used in the case studies. The sparse wording in some prompts is a time and cost-saving measure: in this study, the OpenAI cloud service LLMs were used, which charge by input and output length. The shorter the input, the cheaper the analysis, and if the data is unitized into small units e.g. sentences, then the same prompt needs to be processed over and over again. If there is a lot of data, removing a few words from the prompt can make a difference. This would not be an issue if a local, private LLM would be used, where the only cost is time (though also electricity, which also costs something).

\subsection*{Confirmatory topic classification using LMMs}

This is the classification prompt:

Classify this USSR news Text with a topic Tag. ONLY use one of the Tags defined in this list:
\\ Politics = use for political events and messages, presidiums, Communist Party, Komsomol; communism, socialism, Leninism, political leaders and premiers; foreign governments and politics; international relations; but do NOT use for wars or industry news.
\\ Military = military, national defense, wars, battles; but NOT politics.
\\ Science = scientific and industrial progress and construction; space and aviation; technological advancements and innovations.
\\ Social = social issues and lifestyle, education, students and schools, family, health, leisure, arts, culture, religion, ceremonies.
\\ Disasters = disasters, fires, weather warnings; but NOT wars.
\\ Sports = sports events; sports results and scores; athletic performances.
\\ Farm = domestic USSR agriculture, farming, hunting, kolkhozes; but NOT foreign news, NOT economics.
\\ Industry = USSR economy, domestic business, economic trends, economy plans; industry, mining; industry workers, brigades; but NOT politics, NOT international news, NOT agriculture, NOT tourism.
\\ Misc = use this for any other topics, only if no Tag above fits well; or if topic is unclear.
\\ Classify text: [followed by the synopsis text]

This is the additional exploratory prompt:

You are a cultural historian and Soviet history expert. Analyze this corpus of Soviet era newsreels synopses (one per line, unrelated texts). Come up with a set of the most frequent general content categories that could be used to describe the themes and topics in a newsreels dataset like this. Give me 10 topics, where the last is a Misc topic which subsumes all the remaining less frequent topics. Output the topics as a comma-separated list of keywords or short key phrases, in English. Do not expand the keywords or comment, just give a list. The corpus: \\  ~ [followed by a sample from the corpus]

\subsection*{Historical event and cause detection}

Detect primary first named cause of demise in this ship history Text as either:
\\ Attack = active warfare, bombing, assault, submarine, torpedo; but NOT target practice.
\\ Mine = mine or mine field named as first cause.
\\ Fault = leak, accident, capsizing, engine fail, abandonment, decommission, intentionally sunk, uputatud; but NOT navigation errors like hitting shallows.
\\ Nav = navigation errors like crashing on shallows or reefs; or getting caught in storms, fog, bad weather; but NOT other accidents and NOT intentionally sunk wrecks. 
\\ Text:  [followed by the text entry]

\subsection*{LLMs for missing data augmentation}

Inference prompt for the Finnish-language TV show and film descriptions:

Where is this film or series from? Title: [title]. Description: [description]. \\ 
The info given here is in Finnish because it was shown on YLE but it is not made in Finland. Consider the translated title and synopsis and any actors directors mentioned and your knowledge of TV and guess the production country. NO comments even if unclear, output single country name.

\subsection*{Social network inference from literary texts}

List all pairs of named characters who directly converse in this Text. Format as TSV, no heading. Solve pronoun references, merge references to the same person into name, standardize names. List each pair ONLY once, do NOT repeat listed pairs. Omit people who are just mentioned but do not converse. ONLY list pairs who directly interact. \\ Text: 
[followed by a text segment]

\subsection*{Relevance filtering and OCR correction in digitized newspaper corpora}

Text cleaning prompt:

Fix OCR errors to restore this English newspaper excerpt; do not comment! \\
~ [text]

Classification:

Does this text mention nature or environment in the biological natural world sense, including nature tourism, landscape, agri, environmental policy? Output No if not, or if nature in other senses like natural products, medicine, "human nature" or environment in general non-nature sense. \\ Text: [followed by the text segement]

\subsection*{Text and idea reuse detection}

The synthetic data generation prompts. These consisted of multiple parts:

Targets: \\ always make use of this exact phrase: "Russians are descendants of the Huns" (without quotes). Create examples based on these different ideas but always weave this phrase in, not as a quote but as naturally as possible. \\ and \\
always make some mention Huns or Scythians or Attila or Hunnic culture or language in some way, even if in passing. BUT DO NOT say that Russians could be descendants of the Huns because Russian people are completely unrelated to Huns. Do not even hint that they could be related.'

Prompt:
Write 50 paragraphs drawing from the following fictional alternative timeline ideas. 3 sentences per paragraph. Write some in the style of blogs, some like Reddit, some like forum posts. Assume some paragraphs would be either in the beginning, middle or end of the longer post. All 50 are meant to argue for any combination of these ideas below, but importantly, whatever the topic, [TARGET] \\ Your sample of fictional historical ideas to use: 
\\ Huns originate from the area of present-day Russia.
\\ The West has been fighting an information war for 100+ years against the Russian people by brainwashing. 
\\ Russian education system helps the West brainwash Russians. 
\\ Historical facts (invent some!) have been hidden by those forces.
\\ Russians had an ancient continuous statehood.
\\ Russians had state structures well before Rurik.
\\ Mainstream history considers Rurik as the founder of Russian statehood. 
\\ The first capital of ancient Russia was Slovensk, which ruled a vast territory. 
\\ It was located at the same place as Novgorod and had already been established by 2409 BCE 
\\ Slovensk was founded 3099 years after the creation of the world. 
\\ Orthodox religiousness as a natural part of Russian heritage.
\\ Rus or ancient Russia was called "the Country of Cities" in the old days.
\\ Russians produced all the weapons used in seventeenth-century Europe.
\\ This massive weapons production and export provide proof of a sophisticated civilization.
\\ Now write the 50 paragraphs, single newline separated, DO NOT comment, DO NOT tag the paragraphs, DO NOT number them, also do not forget to mention Huns.

The data distortion prompts (the rephrasing one was done using simple regex):

OCR distortions: Make this text look like it has few OCR errors by splitting couple of words with space and replacing few letters with similar letters.
Russian: Translate into Russian: \\
~[followed by the text]

Classification prompt:

Does this Text mention that Russians are descendants of the Huns or from Hun genetic lineage? Answer Yes if any such ancestry claim. Answer No if no such claim or only talk of cultural or areal links. Text: \\
~[followed by text]

\subsection*{Linguistic usage feature analysis}

Classification prompt; this seemed to work better with the subjective-objective terminology rather than the terms from the paper.

These are modifier-noun phrases from 18th century English texts, a phrase per line. The phrases are unrelated to each other. Output the phrases but append a comma separated value to each indicating if the modifier is Objective or Subjective. \\ Objective tag is for any modifiers describing objective properties such as shape or color or origin (like Persian carpet) or type or material (like wooden chair) and in general attributes that can be defined and measured. Objective also includes fixed compounds like looking glass or eating stall.\\ Subjective tag is for any evaluative modifiers referring to beauty or elegance or usefulness or quality or value or fashion or convenience or rarity. Subjective also includes subjective age (like old car, new house) and subjective size (like spacious house, large garden) and possible users (like men's shirt or family house). Subjective in general is qualitative attributes that cannot be objectively defined.\\
Only guess either Subjective or Objective. The phrase:
\\ ~ [followed by the phrase]

\subsection*{LLM-powered interview analysis}

Data generation prompt:

Write 100 interview responses that in some way discuss the benefits of doing college groupwork assignments online via video call, zoom, messaging apps etc, either as such, or as opposed to live meetings. Invent various positive reasons. Assume the respondent was asked something about online learning in a semi-structured spoken interview. Write like a US college student would but vary response styles. Each response should be short, one sentence but can be a longer sentence. Make it look like they're transcribed speech. Don't use quotation marks.

Data analysis prompt:

Does this student express a Positive or Negative aspect of online groupwork? Student said: [followed by the quote]

\subsection*{Lexical semantic change detection}

Classification prompt; the brackets mark places where the relevant examples were inserted before sending the prompt to the API.

Lexical meaning of "[Target word]" in these two sentences: ignoring minor connotations and modifiers, do they refer to roughly the Same, different but closely Related, distant/figuratively Linked (incl metaphors idioms) or unrelated Distinct objects or concepts? \\
1. [first sentence] \\
2. [second sentence]

\subsection*{Challenging linguistic data annotation in social media posts}

Rate the semantic similarity of the target word in <x> tags in these sentences A and B. Consider only the objects/concepts the word forms refer to: ignore any common etymology and metaphorical similarity! Ignore case! Ignore number (cat/Cats = identical meaning). If target is emoji then rate by its contextual function. Homonyms (like bat the animal vs bat in baseball) count as unrelated. Output numeric rating: 1 is unrelated; 2 is distantly related; 3 is closely related; 4 is identical meaning. \\
A: [first sentence] \\
B: [second sentence] 

\subsection*{ Stance and opinion detection}

Relevance filtering prompt for the police dataset:

Assess if this Text mentions Estonian police or border guard or PPA. Yes = news is from Estonia or about Estonian police or border. If location not hinted assume Estonia. No = news is about another country or police of another country, or context is fictional or metaphorical.
\\ Note this news is titled "[Title]".
\\ Text: [the text]

Stance detection prompt:

Does this Text portray Estonian police or border guard or PPA in Neutral or Positive or Negative light?
\\ Text: [the text]

\subsection*{ Genre detection in literature and film scenes}

This classification prompt works when the token bias is defined in the API call, allowing only the predefined genre labels to be in the output.

Classify genre of this fiction text: [the text segment]

The exploratory book and film script classification prompts:

Classify genre of this fiction Text as either Detective, Fantasy, Sci-Fi, Romance, Thriller, or Other if none of these match. Text: [the text]

Classify genre of this movie Script section as either Detective, Fantasy, Sci-Fi, Romance, Thriller, or Other if none of these match. Script: [the scene from the script]

\subsection*{ A quantitizing approach to literary translation analysis}

Description request prompt; this is after aligning the translation at sentence level with BERTalign:

Determine if there are any significant lexical or stylistic differences between this English text and its Italian translation. If not then output None. If any then explain but only very briefly, do NOT comment otherwise.

The synthetic test set, "semantic edit distance" prompt:

Compare this EN source and JA translation. Ignore minor stylistic differences ignore word order. Focus on objects and subjects and if they differ between EN and JA. Output either: \\ Close translation. \\ Addition = JA contains additional subject object not in EN version. \\ Deletion = something or somebody missing from JA compared to EN. \\ Substitution = only applies if the translation is close but an object or subject in EN has been removed and replaced with something new in JA. \\ Texts: \\ ~[the two texts]

\subsection*{ Zero-shot sense inference for novel words}

The test set generation prompts; there were multiple prompts according to word and language, shown here as a single prompt with bracketed variables:

Write 20 short unrelated sentences which always mention or use this common noun: zoorplick. It's a new word that means [glue, thief, bear], so write about \\
~[so write about glueing and using glue to repair or craft, \\
thieves stealing or robbing, \\
so write about bears in nature] ~ \\
Sentence per line so 20 lines, no numbering no comments. Write in fiction genre, in first or third person, only write grammatical [English, Estonian, Turkish] sentences and in [English, Estonian, Turkish] context.

The inference prompt:

Act as lexicographer skilled in finding and inferring meanings of new words. Based on these Examples, figure out the meaning of "zoorplick", a new noun in  [English, Estonian, Turkish]. Do not comment, just output most likely meaning as a specific single word in English. Examples: \\
~[one or more examples]

\end{document}